\documentclass[10pt,twocolumn,letterpaper]{article}

\usepackage{iccv}
\usepackage{times}
\usepackage[accsupp]{axessibility}  

\usepackage{graphicx}
\usepackage{booktabs}
\usepackage{bm}
\usepackage{multirow}
\usepackage{threeparttable}
\usepackage{epsfig}
\usepackage{algorithm}
\usepackage{algorithmic}
\usepackage{multirow}
\usepackage{threeparttable}
\usepackage{epsfig}
\usepackage{wrapfig}
\usepackage{caption}
\usepackage{lipsum}

\usepackage[breaklinks=true,bookmarks=false]{hyperref}

\usepackage{amssymb}
\usepackage{amsmath}
\usepackage{comment}
\usepackage[capitalize]{cleveref}

\iccvfinalcopy 

\def\iccvPaperID{****} 

\begin{document}


\title{Pluralistic Aging Diffusion Autoencoder}

\author{
Peipei Li$^1$, Rui Wang$^1$, Huaibo Huang$^2$, Ran He$^2$, Zhaofeng He$^1$$^{\star}$\\
$^{1}$Beijing University of Posts and Telecommunications\\
$^{2}$CRIPAC\&MAIS, Institute of Automation, Chinese Academy of Sciences\\
\{lipeipei, wr\_bupt, zhaofenghe\}@bupt.edu.cn, huaibo.huang@cripac.ia.ac.cn,
rhe@nlpr.ia.ac.cn\\
}

\twocolumn[{
\renewcommand\twocolumn[1][]{#1}
\maketitle
\begin{center}
    \captionsetup{type=figure}
    \includegraphics[width=1\textwidth]{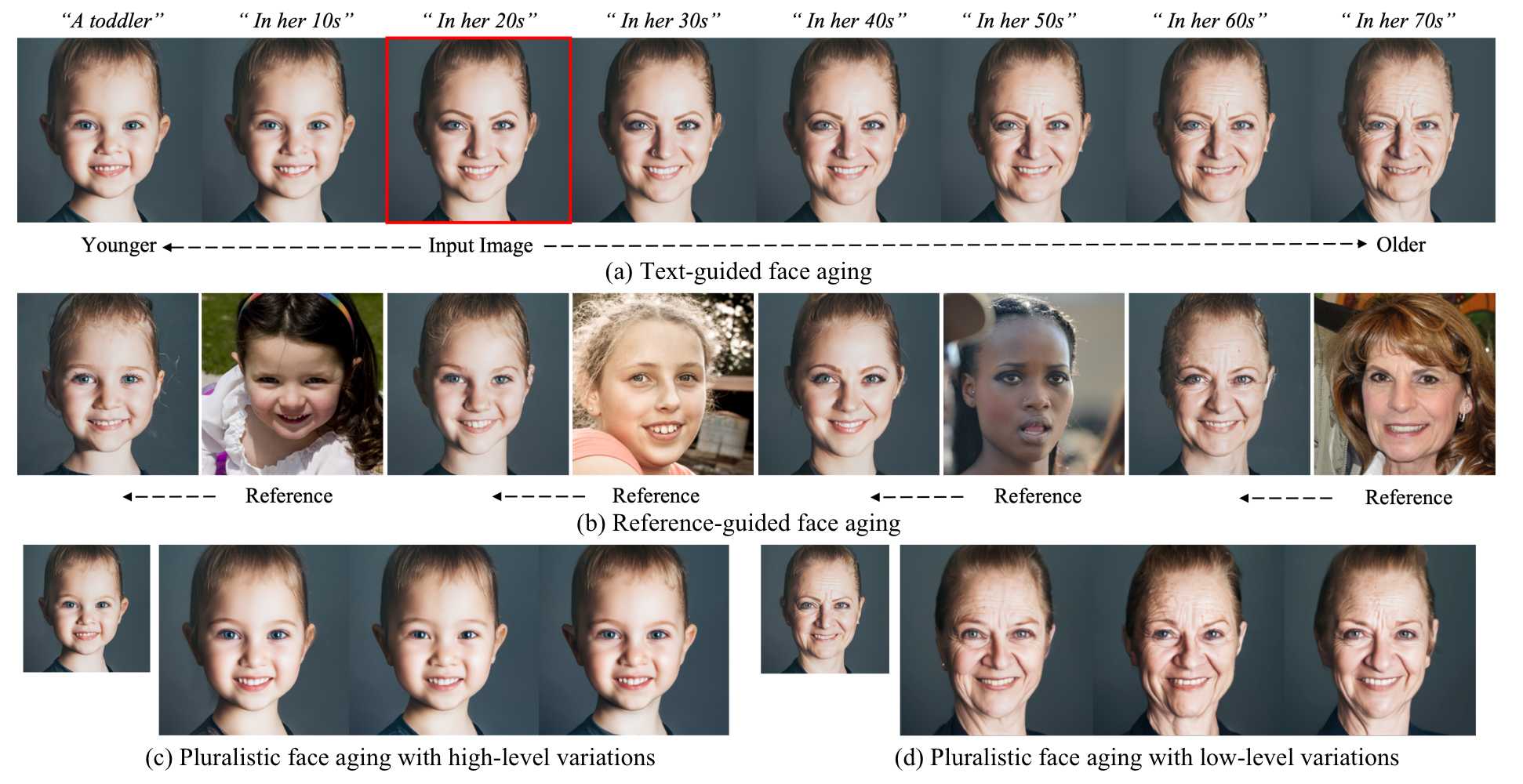}
    \captionof{figure}{Our single framework enables various pluralistic face aging tasks, including (a) Text-guided face aging (the first row), (b) Reference-guided face aging (the second row), (c) Pluralistic face aging with high-level variations (the bottom) and (d) Pluralistic face aging with low-level variations (the bottom).}
    \label{fig:ddim}
\end{center}
}]

 \maketitle
\let\thefootnote\relax\footnotetext{$^{\star}$Corresponding author}
\pagestyle{empty}
\thispagestyle{empty}
\begin{abstract}
    Face aging is an ill-posed problem because multiple plausible aging patterns may correspond to a given input. Most existing methods often produce one deterministic estimation. This paper proposes a novel CLIP-driven Pluralistic Aging Diffusion Autoencoder (PADA) to enhance the diversity of aging patterns. First, we employ diffusion models to generate diverse low-level aging details via a sequential denoising reverse process. Second, 
    we present Probabilistic Aging Embedding (PAE) to capture diverse high-level aging patterns, which represents age information as probabilistic distributions in the common CLIP latent space. A text-guided KL-divergence loss is designed to guide this learning. Our method can achieve pluralistic face aging conditioned on open-world aging texts and arbitrary unseen face images. Qualitative and quantitative experiments demonstrate that our method can generate more diverse and high-quality plausible aging results. 
\end{abstract}

\section{Introduction}
Face aging aims to model facial appearance changes across different ages meanwhile maintaining identity information. It is an ill-posed learning problem due to multiple plausible aging results for a given input. Given various aging images in the last row of Fig.~\ref{fig:ddim}, which one meets your imagination of aging? Since human aging process is influenced by a variety of factors, including genetics and social environment, there may be significant differences in both general aging trends and local details. Pluralistic face aging aims to generate multiple and diverse plausible face aging images from a single input.

Deep generative models, such as generative adversarial networks (GANs)~\cite{goodfellow2014generative} and variational autoencoders (VAEs)~\cite{kingma2019introduction}, have shown impressive performance in terms of face aging~\cite{lihierarchical,or2020lifespan,makhmudkhujaev2021re,alaluf2021only,he2021disentangled}. Unfortunately, most previous methods can only produce one ``optimal'' aging pattern, which is inconsistent with human cognition.
Recently, diffusion models~\cite{dhariwal2021diffusion,kingma2021variational} show comparable or even better generation quality compared to GANs, which learn the reverse of a particular Markov diffusion process and cover the modes of data distribution better. 
Inspired by this, we intend to employ diffusion models to generate aging faces with low-level subtle stochastic variations, such as diverse 
wrinkles, as shown in the lower right of Fig.~\ref{fig:ddim} (d).

\begin{figure}[t]
\begin{center}
\includegraphics[width=1\linewidth]{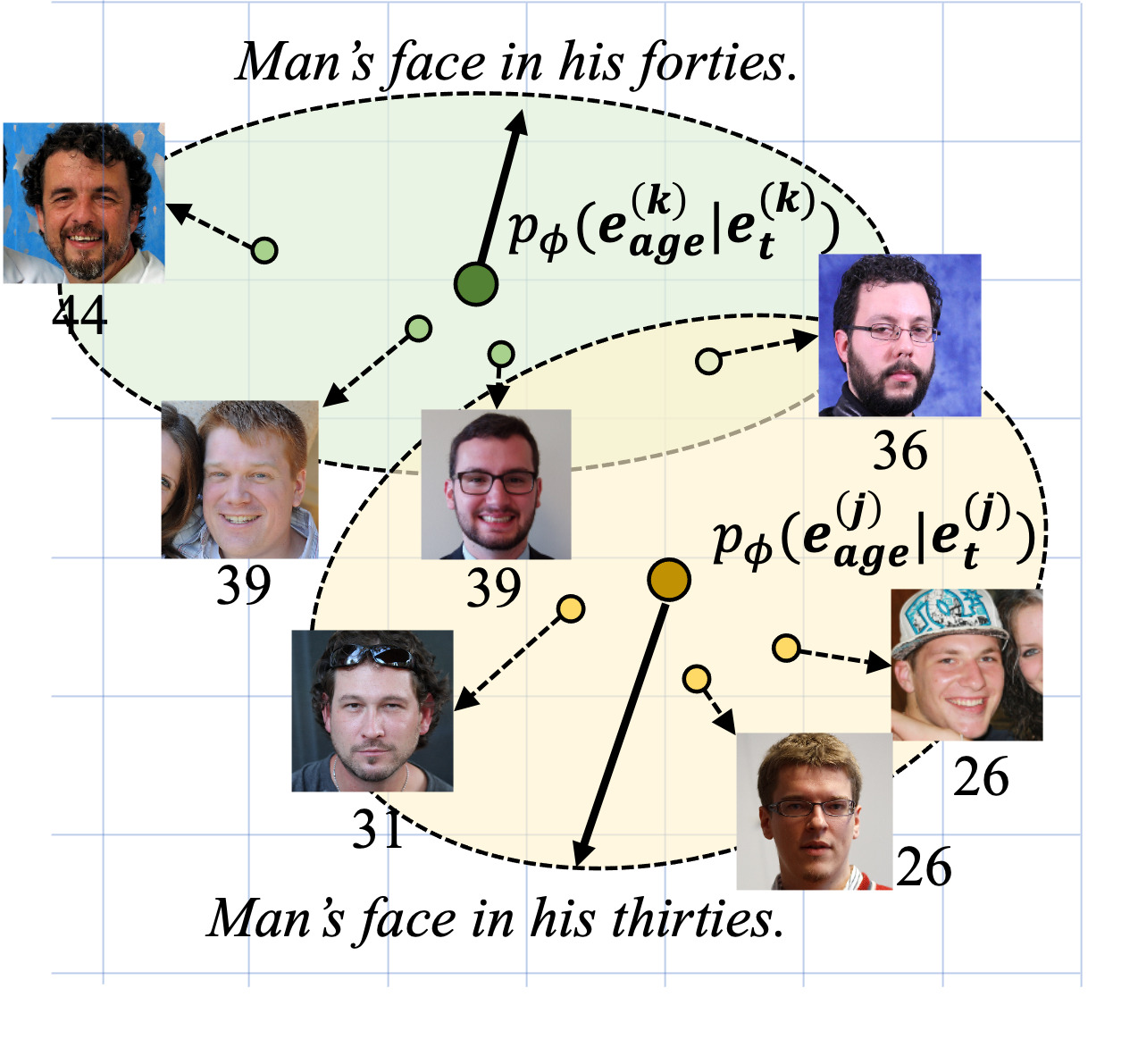}
\end{center}
   \caption{One-to-many correspondences between coarse text-based age feature and image-based age features in the CLIP latent space.
   The solid line indicates the text-based age feature while the dashed line indicates the image-based age features.
   }
\label{fig:analysis}
\end{figure}

In addition to low-level stochastic details, the aging process is accompanied by high-level age semantic changes, such as, getting fatter or thinner, getting darker or whiter, as shown in Fig.~\ref{fig:ddim} (c). Previous face aging methods~\cite{or2020lifespan,alaluf2021only,he2021disentangled,gomez2022custom} 
directly represent the target age as a deterministic point or direction in the latent space, ignoring the personalized age characteristics.
So here comes a key challenge for pluralistic face aging: how to learn high-level age representations with stochastic variations. 
To address it, we draw support from the pre-trained  CLIP~\cite{radford2021learning} model and propose Probabilistic Aging Embedding (PAE), which represents age information as a distribution rather than a deterministic point.
The intuition to leverage CLIP is illustrated in Fig.~\ref{fig:analysis}. In the well-aligned image-text latent space, there are likely to be multiple image-based age features for a coarse text-based age feature of \textit{``Man's face in his forties"}. 
Inspired by it, we attempt to model PAE in CLIP latent space to capture the stochastic high-level age semantics.

In this paper, we explore pluralistic face aging based on both text and image conditions. We propose a CLIP-driven Pluralistic Aging Diffusion Autoencoder (PADA) to simultaneously model low-level stochastic variations and high-level age semantic variations in the aging process. 
For the low-level age variations, our method is based on the diffusion model, which can generate stochastic low-level face details via a sequential denoising procedure. 
For the high-level age variations, we propose Probabilistic Aging Embedding (PAE) by representing the age information as probabilistic distributions in the CLIP space. Specifically, we represent the age information as a multivariate Gaussian distribution rather than a deterministic point, where the mean of the distribution indicates average age information, while the variance indicates the personalized aging patterns of the image.
Then we feed back the PAE into the diffusion model with an adaptive modulation for pluralistic face aging.
Since our goal is to learn the diverse aging patterns and achieve face aging with preservation of age-irrelevant information (i.e., identity and background), three types of losses are employed: 1) Text-guided KL-divergence loss; 2) Age fidelity loss; 3) Preservation loss.
To summarize, our contributions are four-fold:
\begin{itemize}
\item We propose a novel CLIP-driven Pluralistic Aging Diffusion Autoencoder (PADA) for pluralistic face aging, which can generate diverse aging results with both high-level age semantic variations and low-level stochastic variations.

\item Probabilistic Aging Embedding (PAE) is proposed in the common CLIP space to represent the diverse high-level aging patterns as probabilistic distribution, where a text-guided KL-divergence loss is employed to guide this learning.

\item A more user-friendly interaction way for face aging is provided, which can achieve age manipulation conditioned on both the open-world age descriptions and arbitrary unseen face images in the wild. 

\item Extensive qualitative and quantitative experiments show that our method outperforms the state-of-the-art aging methods and can generate plausible and diverse aging patterns. 

\end{itemize}

\begin{figure*}
\begin{center}
\includegraphics[width=1\linewidth]{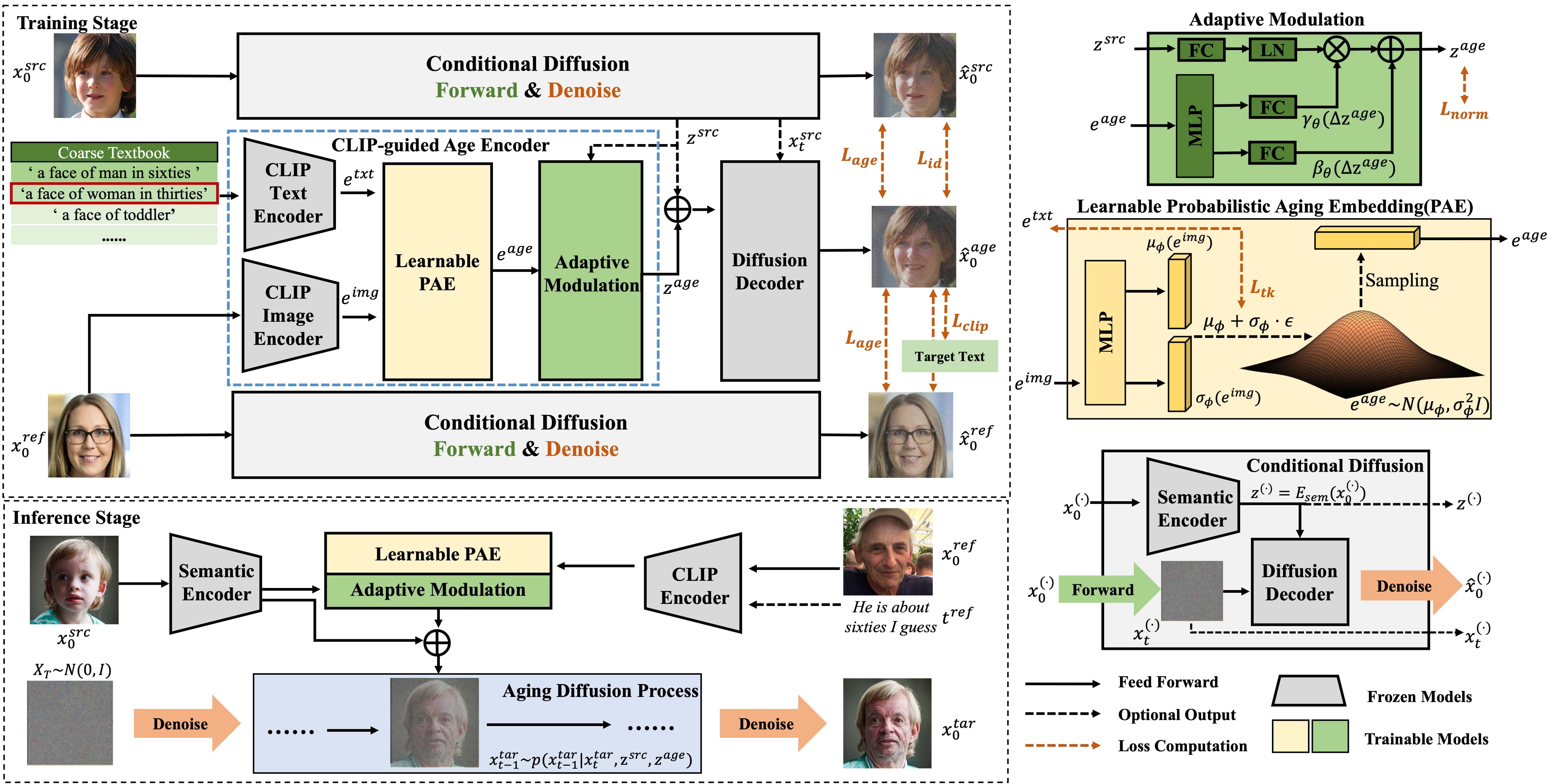}
   \caption{
Overview of the proposed PADA, which consists of a conditional DDIM decoder, a semantic encoder, and a CLIP-guided age encoder. 
The semantic information $z^{src}$ is extracted from the input image $x_{0}^{src}$ via the semantic encoder. Meanwhile, the Probabilistic Aging Embedding (PAE) $e^{age}$ is obtained by the CLIP-guided age encoder and translated to $z^{age}$ in an adaptive manner. Based on $z^{src}$ and $z^{age}$, we can generate pluralistic aging results via the conditional DDIM decoder. In the inference stage, $e^{age}$ can also be sampled from text-based age prior.
}
\label{fig:framwork}
\end{center}
\end{figure*}

\section{Related Work}

\textbf{Face Aging} is one of the most challenging components in modern face manipulation, which is an ill-posed problem with multiple aging results corresponding to the same input. Generative Adversarial Network (GAN)~\cite{goodfellow2014generative} has been successfully applied to face aging and made impressive results~\cite{gomez2022custom,li2019global,li2018global,lihierarchical,makhmudkhujaev2021re,liu2019attribute,li2020deep,or2020lifespan,alaluf2021only,he2021disentangled}. 
LATS~\cite{or2020lifespan} successfully models both the texture transformation and shape deformation. SAM~\cite{alaluf2021only} is based on a pre-trained StyleGAN~\cite{karras2019style} and can generate high-quality aging results. 
RAGAN~\cite{makhmudkhujaev2021re} proposes a personalized self-guidance module, which leverages the interactions between identity and target age to learn the personalized age features. CUSP~\cite{gomez2022custom} disentangles the style and content of the input, providing structure modifications with relevant details unchanged. 
Variational Autoencoder (VAE) shows promising ability in generating results with interpretability~\cite{kingma2019introduction} and is employed for face aging~\cite{zhang2017age,lihierarchical}. Li et al.~\cite{lihierarchical} propose a disentangled adversarial autoencoder (DAAE)
for face aging, which disentangles the images into three independent factors, including age, identity and extraneous information. 
All in all, although there may be many reasonable possibilities, the most existing methods produce only one ``optimal'' estimation for each input.
Inspired by probabilistic embedding learning~\cite{lee2020uncertainty,debnath2021uac,chun2021probabilistic} , we intend to utilize pre-trained CLIP to model Probabilistic Aging Embedding (PAE) when paired aging data is unavailable.

Recently, \textbf{Diffusion Probabilistic Models (DPM)} have achieved 
state-of-the-art results in generation quality as well as in sample quality~\cite{dhariwal2021diffusion,kingma2021variational,nichol2021improved, kim2021diffusionclip}, which consist of a forward (or inference) Markovian diffusion process $q$ and a learned reverse (or generative) diffusion process $p_{\theta}$:$
q\left ( x_{1:T}|x_{0} \right )=\prod_{t=1}^{T}q\left ( x_{t}|x_{t-1}\right ),
p_{\theta} \left ( x_{0:T} \right )=p\left ( x_{T} \right )\prod_{t=1}^{T}p_{\theta}\left ( x_{t-1}|x_{t} \right ).
$
More formally, we define the forward diffusion process at time $t$ by gradually adding Gaussian noise to the input $x_{0}$ as $q\left ( x_{t}|x_{t-1} \right )=\mathcal{N}\left ( x_t|\sqrt{\alpha _t}x_{t-1},\left ( 1-\alpha _t \right )I \right )$; while the inverse process by gradually removing the noise from the Gaussian noise $p_{\theta} \left ( x_{t-1}|x_t \right )=\mathcal{N}\left ( x_{t-1}| \mu _{\theta }\left ( x_t ,t\right ),\sigma _{t}^{2}I\right ).$
Thus, DDPMs require simulating a Markov chain with many iterations to produce a high quality sample, which limits its generation efficiency.
Denoising Diffusion Implicit Models (DDIMs)~\cite{song2020denoising} are then proposed to accelerate sampling, which share the same training objective function with DDPMs, but employ non-Markovian diffusion processes. The deterministic generative process and the following inference distribution are shown as follows:
\begin{equation}
\label{eq:1}
\begin{array}{c}
x_{t-1}=\sqrt{\alpha_{t-1}}f_\theta(x_t,t)+\sqrt{1-\alpha _{t-1}}\epsilon _{\theta }^{t}\left ( x_{t} \right ),
\end{array}
\end{equation}
\begin{equation}
\label{eq:2}
\begin{array}{c}
q( x_{t-1}|x_{t}, x_{0})=\mathcal{N}\left(\sqrt{{\overline{\alpha _{t-1}}}}x_{0}+\sqrt{1-{\overline{\alpha_{t-1}}}}\frac{x_{t}-\sqrt{\overline{\alpha_{t}}}x_{0}}{\sqrt{1-\overline{\alpha_{t}}}},\bm{0} \right ),
\end{array}
\end{equation}
where $f_\theta(x_t,t)$ is the prediction of $x_0$ at timestep $t$ given predicted noise $\epsilon _{\theta }^{t}(x_t)$ :
\begin{equation}
\label{eq:3}
\begin{array}{c}
f_\theta(x_t,t) = \frac{{x_{t}-\sqrt{1-\overline{\alpha_{_t}}}\epsilon _{\theta }^{t}(x_{t})}}{\sqrt{\overline{\alpha _{t}}}}.
\end{array}
\end{equation}
To learn a meaningful latent space, DiffAE~\cite{preechakul2022diffusion} trains DDPM with an extra encoder, which embeds the input into a latent vector to guide the reverse diffusion process.

With the development of the powerful cross-modal visual and language model CLIP~\cite{radford2021learning}, many recent efforts start to study \textbf{CLIP-Guided Image Generation}~\cite{patashnik2021styleclip, kwon2022clipstyler, sun2022anyface, wei2021hairclip, gal2022stylegan, kim2021diffusionclip, chen2023vlp}.
CLIPStyler~\cite{kwon2022clipstyler} utilizes the well-aligned CLIP latent space for high-quality style transfer. 
HairCLIP~\cite{wei2021hairclip} is based on pre-trained StyleGAN~\cite{karras2020analyzing} and CLIP for a more user-friendly design of hairstyles. 
However, there is no existing method fully leverage the CLIP latent space for lifespan aging. In this paper, we explore CLIP latent space to provide diverse age representations for face aging. 

\section{Proposed Method}
\subsection{Overview}

We propose CLIP-driven Pluralistic Aging Diffusion Autoencoder (PADA), a conditional DDIM to generate multiple face aging results. Specifically, we aim to transform the source image $x_0^{src}$ to a set of target aging results $\{(x_0^{tar})^{(i)}\}_{i=1}^N$ conditioned on either reference image ${x^{ref}_0}$ or text description ${t}^{ref}$. $x_0^{tar}$ is the target aging result with stochastic age variations, which is generated via an aging reverse process of DDIM:
\begin{equation}
\label{eq:reverse}
\begin{array}{c}
p(x^{tar}_{0:T}|z^{src}, z^{age})=p(x_T)\prod \limits_{t=1}^T p(x^{tar}_{t-1}|x^{tar}_t,z^{src},z^{age}).
\end{array}
\end{equation}
Specifically, our aging reverse process is based on three parts, including a pre-trained conditional DDIM decoder $p(x^{tar}_{t-1}|x^{tar}_t,z^{src},z^{age})$, a pre-trained semantic encoder $z^{src}=E_{sem}(x^{src}_0)$, and a CLIP-guided age encoder $z^{age}=E_{age}(x_0^{ref}, t^{ref})$, where $z^{src}$ is the semantic information of input image $x_0^{src}$; and $z^{age}$ is the stochastic age condition learned from reference image $x_0^{ref}$ or text $t^{ref}$.
An overview of our architecture is shown in Fig.~ \ref{fig:framwork}.

Unlike previous methods~\cite{or2020lifespan,alaluf2021only,he2021disentangled, gomez2022custom} that learn a determined age variable, we propose CLIP-guided age encoder $E_{age}$ to learn the stochastic age condition $z^{age}$ for pluralistic face aging. 
Concretely, leveraging the well-aligned text-image latent space of the pre-trained CLIP model, we extract the \textbf{Probabilistic Aging Embedding (PAE)}.
Then, an adaptive modulation mechanism is utilized to translate PAE into the stochastic age condition $z^{age}$. Conditioned on both $z^{age}$ and $z^{src}$, we can generate pluralistic aging results via the pre-trained conditional DDIM decoder. The training and inference algorithms are detailed in the supplementary materials.

\subsection{Stochastic Age Condition} 
\label{sec:subsec32}

In this section, we introduce two key ingredients for the learning of stochastic age condition $z^{age}$: Probabilistic Aging Embedding (PAE) and Adaptive Modulation.

\subsubsection{Probabilistic Aging Embedding} 
As illustrated in Fig.~\ref{fig:analysis}, in the well-aligned image-text latent space of the pre-trained CLIP~\cite{radford2021learning} model,
the cosine similarity of related images and texts is maximized, while that of unrelated images and texts is minimized. 
Since coarse text-based age features contain the average aging information (e.g., \textit{Man's face in his forties}), rich personalized age-related fine features are assumed to be most likely distributed around it.
Therefore, to obtain richer personalized age-related features, we propose Probabilistic Aging Embedding (PAE) based on well-aligned CLIP latent space.

Concretely, we represent our PAE as a multivariate Gaussian distribution $\mathcal{N}(e^{age};\mu_\phi, \sigma^2_\phi I)$ in the CLIP latent space.
The mean $\mu_\phi$ indicates average age information, while the variance $\sigma_\phi$ indicates the personalized aging patterns from the reference image. 
We assume the posterior approximation $p_\phi(e^{age}|e^{img})$ follows an isotropic multivariate Gaussian:
\begin{equation}
\label{eq:1}
\begin{array}{c}
p_\phi(e^{age}|e^{img})=\mathcal{N}(e^{age}; \mu_{\phi}(e^{img}), \sigma^{2}_{\phi}(e^{img}) I),
\end{array}
\end{equation}
where $e^{img}$ is extracted from the reference image $x_{0}^{ref}$ via the CLIP image encoder. $\mu_{\phi}(\cdot)$ and $\sigma_{\phi}(\cdot)$ are implemented with a shared MLP-backbone and two head branches. 
Moreover, we introduce an age prior for PAE:
$p(e^{age})=\mathcal{N}(e^{txt}, I)$,
where $e^{txt}$ is extracted from the coarse age description $t^{ref}$ via the CLIP text encoder.

Correspondingly, we use a text-guided KL-divergence loss for prior matching. 
The learned probabilistic aging embedding $e^{age}$ is sampled via a reparameterization trick:
$e^{age}=\mu_{\phi}+ \sigma_{\phi}\odot \eta$,
where $\eta\sim\mathcal{N}(0, I)$ and $\odot$ refers to element-wise multiplication. 
To get richer aging features, we also directly model PAE in a text-driven manner:
$e^{age}=e^{txt}+\epsilon\cdot \eta$, where $\epsilon$ is a hyperparameter for sampling intensity.
We can prove that by careful selection of $\epsilon$, probabilistic aging embeddings $e^{age}$ can be reliably sampled on the hyper-sphere of CLIP latent space. The formal proof is provided in the supplementary materials.

In this way, we can sample probabilistic aging embedding $e^{age}$ from our learned distribution.

\subsubsection{Adaptive Modulation}
Our Adaptive Modulation is designed to translate the learned $e^{age}$ into stochastic age condition $z^{age}$, which is composed of an MLP-based backbone and a modulation module.  
Concretely, for each of the sampled $e^{age}$, we first translate it into semantic latent space of the conditional DDIM: $\Delta z^{age}=MLP(e^{age})$.
Then the aging information $\Delta z^{age}$ is transferred into the stochastic age condition $z^{age}$ by our modulation module, which can be formulated as:
\begin{equation}
\label{eq:fusion}
\begin{aligned}
z^{age} = \gamma_\theta(\Delta z^{age})\frac{z^{src}-\mu^{src}}{\sigma^{src}} + \beta_\theta(\Delta z^{age}),
\end{aligned}
\end{equation}
where $\mu^{src}$ and $\sigma^{src}$ are the channel-wise mean and standard deviation of $z^{src}$, respectively. Here, we construct $\gamma_\theta(\cdot)$ and $\beta_\theta(\cdot)$ with 2 fully-connected layers.
Conditioned on both $z^{src}$ and $z^{age}$, we can generate pluralistic face aging results $x_0^{tar}$ via the pre-trained DDIM decoder $p(x^{tar}_{t-1}|x^{tar}_t,z^{src},z^{age})$.

\subsection{Loss Functions}
Our goal is to  achieve pluralistic face aging with age-irrelevant information well-preserved. Therefore, three types of loss objectives are designed, including text-guided KL-divergence loss, age fidelity 
loss, and preservation loss. 
To avoid confusion, here we denote $\hat{x}^{(\cdot)}_0$ as the approximate reconstruction of ${x^{(\cdot)}_0}$ at timestep $t$ following Eq.(\ref{eq:3}). For example,  $\hat{x}^{tar}_0$ means the approximate reconstruction of aging result $x^{tar}_0$ at time step $t$.

\textbf{Text-guided KL-divergence Loss.} 
To prevent the learned variances from collapsing to zero, we explicitly constrain PAE to be close to a Gaussian distribution 
by introducing a text-guided KL-divergence loss $L_{tKL}$:

\begin{equation}
\label{eq:1}
\begin{aligned}
 L_{tKL} &= D_{KL}\left(\mathcal{N}(e^{age};\mu_\phi,\sigma_\phi^2 I) || \mathcal{N}(e^{txt}, I)\right)\\
&= \frac{1}{2}\sum_j^D\left((\mu_\phi^j-(e^{txt})^j)^2 - 1 + (\sigma_\phi^j)^2 - log(\sigma_\phi^j)^2\right),
\end{aligned}
\end{equation}
where $D$ denotes the channel of $e^{age}$.
For training stability, we replace the Euclidean distance term
with a cosine similarity term $-cos(\mu_\phi,e^{txt})$ and a norm term $||\mu_\phi-e^{img}||_2$, which is optimally equivalent when considering CLIP spaces as hyperspheres \footnote{In practice, we normalize the features in CLIP latent space by their $L_2$ norm.}. More details can be found in supplementary materials.

\textbf{Age Fidelity Loss.} To better maintain age fidelity, we employ an age prediction loss and a clip directional loss. Concretely, we first use a pre-trained age estimator DEX~\cite{rothe2018deep} to estimate the age of intermediate generation $\hat{x}_0^{tar}$ in Eq.(\ref{eq:3}). Due to the blurry of $\hat{x}_0^{tar}$, we use an aging triplet loss for better age fidelity. For conciseness, we denote $f^{(\cdot)}$ as aging representation in the last fully-connected layer of the age predictor:
\begin{equation}
\label{eq:age_tri}
\begin{aligned}
L_{age}=\max\left\{ \langle{f^{src},f^{tar}}\rangle-\langle{f^{ref},f^{tar}}\rangle+m, 0\right\},
\end{aligned}
\end{equation}
where ${\langle{\cdot, \cdot}\rangle}$ refers to the cosine similarity and $m$ refers to the margin. 

To learn more aging details, we additionally employ a CLIP directional loss $L_{clip}$ \cite{patashnik2021styleclip}.

\textbf{Preservation Loss.} To better preserve the identity and age-irrelevant information, we employ identity loss $L_{id}$, norm loss $L_{norm}$, and reconstruction loss $L_{rec}$. 
Concretely, the identity preservation loss $L_{id}$ is used to preserve the identity during generation, which is formulated as:
    \begin{equation}
    \label{eq:norm}
    \begin{array}{c}
    L_{id} = -cos(R(\hat{x}_0^{src}), R(\hat{x}_0^{tar})),
    \end{array}
    \end{equation}
where $R(\cdot)$ is output of the final fully-connected layer of the pre-trained ArcFace~\cite{deng2019arcface}.
Norm loss $L_{norm}$ is the regularization term to ensure the generation quality: 
    \begin{equation}
    \label{eq:norm}
    \begin{array}{c}
    L_{norm} = ||z^{age}||_2^2.
    \end{array}
    \end{equation}
To ensure the age-irrelevant information unchanged, we introduce the reconstruction loss $L_{rec}$, which is defined as:
    \begin{equation}
    \label{eq:rec}
    \begin{array}{c}
    L_{rec} = ||\hat{x}_0^{src} - \hat{x}_0^{tar}||_2^2.
    \end{array}
    \end{equation}
    The overall loss of PADA is formulated as:
    \begin{equation}
    \begin{aligned}
    L = L_{age} + \lambda_1 L_{clip} &+\lambda_2 L_{tKL}+\lambda_3 L_{id} \\ 
    &+ \lambda_4 L_{norm} + \lambda_5 L_{rec},
    \end{aligned}
    \end{equation}
where $\lambda_i$ is hyperparameter controlling the weight of each loss. The details are described in Section \ref{sec: Implementation}.

\begin{figure*}
\begin{center}
\includegraphics[width=1\linewidth]{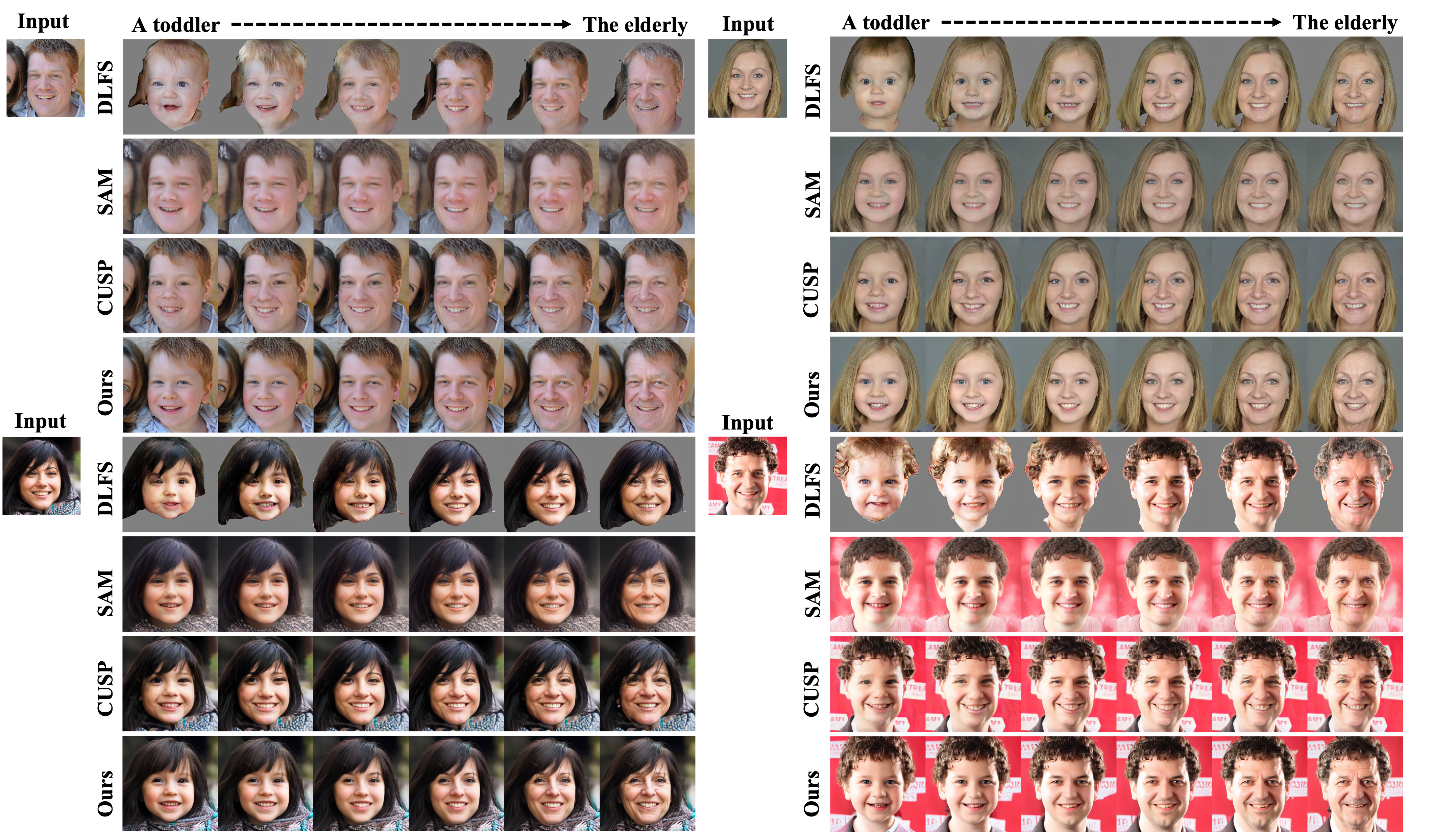}
\end{center}
   \caption{Qualitative comparison with DLFS~\cite{he2021disentangled}, SAM~\cite{alaluf2021only}, and CUSP~\cite{gomez2022custom} on the FFHQ-AT test set. Best viewed zoomed-in.}
\label{fig:ffhq}
\end{figure*}

\section{Experiments}
\textbf{Datasets.}
We relabeled a new facial aging dataset based on the images of FFHQ~\cite{karras2019style}, FFHQ-AT. We roughly divided the FFHQ into 7 age groups: \textit{0-5, 6-15, 16-25, 26-35, 36-50, 51-70, 70+} and accordingly pre-defined 14 age-related descriptions, including \textit{`a toddler girl's/boy's face',  `woman's/man's face in her/his tens', `woman's/man's face in her/his twenties', `woman's/man's face in her/his thirties', `woman's/man's face in her/his forties', `woman's/man's face in her/his sixties', and `woman's/man's face in her/his eighties'}. Then, we use a pre-trained DEX\footnote{https://github.com/siriusdemon/pytorch-DEX} to predict the age value of the image in FFHQ. According to the 
absolute difference value between the predicted age and the pre-defined age, we annotate the image with the pre-defined text description. 
Note that, with these simple text descriptions, our PADA can achieve face aging based on arbitrary age-related descriptions, such as \textit{`a face of teenager'}.
We choose 66,928 images as training set and 3,072 images as test set, which contains 1,528 males and 1,544 females. Finally, we train our model on the new relabeled FFHQ-AT and evaluate it on both FFHQ-AT and CelebA-HQ~\cite{lee2020maskgan} test sets.

\begin{figure*}
\begin{center}
\includegraphics[width=1\linewidth]{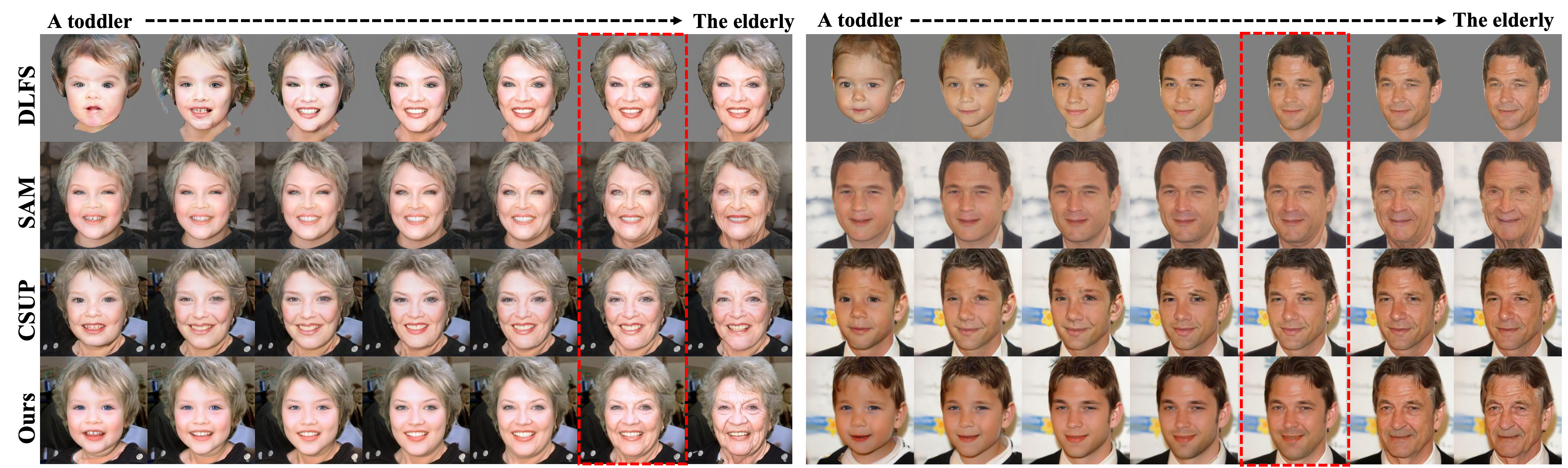}
\end{center}
   \caption{Generalization ability and comparison with DLFS~\cite{he2021disentangled}, SAM~\cite{alaluf2021only}, and CUSP~\cite{gomez2022custom} on the CelebA-HQ test set. 
   }
\label{fig:celeba}
\end{figure*}

\textbf{Implementation Details.} 
\label{sec: Implementation}
The conditional DDIM decoder and semantic encoder~\cite{preechakul2022diffusion} are pre-trained on the FFHQ dataset \cite{karras2019style}. 
In our $L_{rec}$, we randomly set the reference images equal to the source images with the probability of 0.167.
Aging results are generated with $T=25$ steps in all cases. 
Our implementations are based on the MindSpore.
During training, we choose our hyperparameters with $\lambda_1=0.6$, $\lambda_2=0.01$, $\lambda_3=0.2$, $\lambda_4=0.01$, and $\lambda_5=0.1$. We set the margin $m$ in $L_{age}$ as 0.15 and the sampling intensity $\epsilon$ as 0.001, respectively. The Adam optimizer~\cite{kingma2014adam} is used with $lr=0.0001$, $\beta_1=0.9$, $\beta_2=0.999$. PADA is trained for 20 epochs with a batch size of 24.

\textbf{Evaluation Metrics.} We automatically and manually compare our method with three state-of-the-art aging methods, including DLFS\cite{he2021disentangled}, SAM\cite{alaluf2021only}, and CUSP\cite{gomez2022custom}.  We compare them from three aspects: aging accuracy, identity preservation, and aging quality. 
For automatically evaluation, 
1) since the pre-trained DEX~\cite{rothe2018deep} is used for training, we employ Face++ API \footnote{Face++ Face detection API: https://www.faceplusplus.com.} for aging accuracy (Age MAE) evaluation; 
2) since the pre-trained Arcface~\cite{deng2019arcface} is used for training, we employ the pre-trained SFace~\cite{zhong2021sface} for identity preservation evaluation; 
3) for the aging quality evaluation, 
we employ Fréchet Inception Distance (FID) ~\cite{heusel2017gans} to assess the discrepancy between the generated images and the real ones with the same age. 
For manual evaluation, we perform a human evaluation to reliably compare the performance with different aging methods~\cite{he2021disentangled,alaluf2021only,gomez2022custom}.
Besides, we conduct the ablation study and explore more interesting properties of PADA, including diversity exploration, text-guided, and reference-guided face aging.

\subsection{Comparison with Face Aging Methods}
We compare our PADA with three state-of-the-art face aging methods: DLFS~\cite{he2021disentangled}, SAM~\cite{alaluf2021only}, and CUSP~\cite{gomez2022custom}. For all these methods, we employ their official implementation and pre-trained models. Note that the above methods are capable of covering different ranges of aging. Therefore, following ~\cite{gomez2022custom},
we compare the age group-based generation results with the three state-of-the-art methods.

\textbf{Qualitative Comparison.} 
We show the face aging results from 0 to 70 years old with 10-20 age intervals for qualitative comparison on FFHQ-AT test set.
Following DLFS~\cite{he2021disentangled}, we generate missing aging images by interpolation in the latent space. As shown in Fig.~\ref{fig:ffhq}, our PADA outperforms the three state-of-the-art face aging methods DLFS~\cite{he2021disentangled}, SAM~\cite{alaluf2021only} and CUSP~\cite{gomez2022custom} in terms of shape deformation, texture transformation and generation quality. 
For shape deformation, our PADA and DLFS~\cite{he2021disentangled} can generate plausible baby faces with round shape and short baby teeth, while SAM~\cite{alaluf2021only} and CUSP~\cite{gomez2022custom} fail at shape deformation modeling, especially for babies. However, DLFS~\cite{he2021disentangled} tends to produce pronounced artifacts, resulting in less reliability. For texture transformation, our PADA and SAM~\cite{alaluf2021only} can produce 
detailed textures and achieve neck aging. 
CUSP~\cite{gomez2022custom} and DLFS~\cite{he2021disentangled} may generate artifacts on the teeth and facial contours. For generation quality, both our PADA and CUSP~\cite{gomez2022custom} can preserve pretty background information. Meanwhile, due to the GAN inversion~\cite{richardson2021encoding}, SAM~\cite{alaluf2021only} cannot accurately project real images into latent space, leading to blurred background. DLFS~\cite{he2021disentangled} can only generate facial images without background.

We also compare with the three state-of-the-art face aging methods DLFS~\cite{he2021disentangled}, SAM~\cite{alaluf2021only}, and CUSP~\cite{gomez2022custom} on CelebA-HQ~\cite{lee2020maskgan} test set in Fig.~\ref{fig:celeba}. Obviously, our PADA outperforms these methods on both shape deformation and texture transformation. Besides, our PADA can generate geriatric spots (the right subject in Fig.~\ref{fig:celeba}), which cannot be achieved by other methods.

We also compare our PADA wit DiffAE~\cite{preechakul2022diffusion}, which is a strong diffusion-based model for image manipulation. Here, we employ the official implementation of DiffAE~\cite{preechakul2022diffusion} as baseline for face aging.
As shown in Fig.~\ref{fig:agespace}, since DiffAE~\cite{preechakul2022diffusion} ($z^{src}$ manipulation) achieves face aging by linear interpolation in the latent space, it ignores the non-linearity of aging process and cannot generate images of specified ages. PADA ($z^{age}$ manipulation) achieves more reliable aging results with complex non-linear aging directions. Moreover, we incorporate CLIP encoders for flexible face aging and PAE for pluralistic aging with high-level variances.

\begin{figure*}[h]
\centering
\includegraphics[width=1\linewidth]{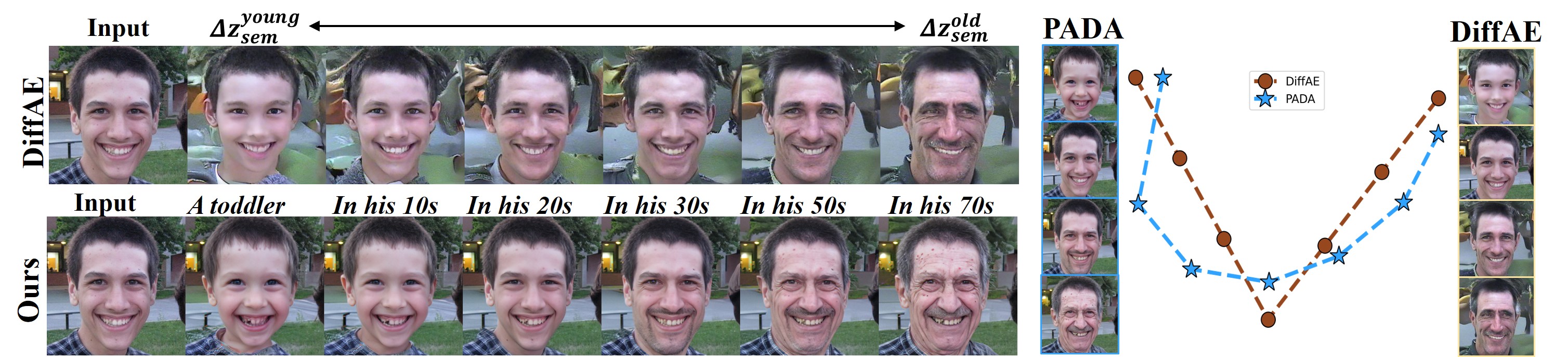}
   \caption{Compared with DiffAE. To compare the learned aging directions of DiffAE and PADA, we use PCA to project the learned latent codes obtained from 14 different images (right). Zoomed-in.}
\label{fig:agespace}
\end{figure*}

\textbf{Quantitative Comparison.} Both aging accuracy and identity preservation are essential quantitative metrics for face aging. We report the quantitative comparison results with DLFS~\cite{he2021disentangled}, SAM~\cite{alaluf2021only}, and CUSP~\cite{gomez2022custom} in Table~\ref{tab:quan}.
As expected, our PADA achieves the best performance on both the identity preservation and aging accuracy (Age MAE).  Besides, our PADA also achieves best performance on aging quality (FID~\cite{heusel2017gans}), which supports the qualitative analysis in Fig.~\ref{fig:ffhq} and Fig.~\ref{fig:celeba}.

\begin{table}\footnotesize
\centering
\caption{Quantitative comparison on FFHQ-AT test set.}
\label{tab:quan}
\renewcommand{\arraystretch}{1} 
\begin{tabular}{c|c|c|c}
  \toprule
  & Age MAE ($\downarrow$) & ID Preservation ($\uparrow$) &FID ($\downarrow$)\\
  \hline
  \multirow{1}*{DLFS\cite{he2021disentangled}}& 10.36 & {0.5421} & 73.98\\ 
  
   \multirow{1}*{SAM\cite{alaluf2021only}}& 9.31 & 0.5161 & 56.86\\
   
  \multirow{1}*{CUSP\cite{gomez2022custom}}& 12.3 & 0.5320 & {37.01}\\
  \hline
  \multirow{1}*{Ours} & \textbf{9.19} & \textbf{0.6516} & \textbf{16.71} \\
  \hline
  \end{tabular}
\end{table}

\begin{table}\scriptsize
\centering

\caption{Human evaluation on FFHQ-AT test set. Overall represents the results of all three age groups.}
\label{tab:user}
\renewcommand{\arraystretch}{1.1} 
 \setlength\tabcolsep{3pt}
\begin{tabular}{rcccc|c}
  \toprule
  && DLFS \cite{he2021disentangled}  & SAM \cite{alaluf2021only}& CUSP \cite{gomez2022custom}& Ours \\
  \hline
  \multirow{4}*{Aging Accuracy ($\uparrow$)}& \textbf{Overall}& 32.33&38.87&37.86&\textbf{46.29}\\
& \textit{`a toddler'} & \textbf{46.67}&37.05&32.95&46.41\\
& \textit{`an adult'}&24.87&41.28&41.15&\textbf{52.31}\\
&\textit{`the elderly'}&24.62&41.28&44.10&\textbf{50.64}\\
  \hline
 \multirow{4}*{ID Preservation ($\uparrow$)} & \textbf{Overall}&34.91&40.68&32.78&\textbf{40.80}\\
& \textit{`a toddler'}&56.05&\textbf{57.41}&54.15&56.73\\
& \textit{`an adult'}&54.97&55.10&54.28&\textbf{56.19}\\
&\textit{`the elderly'}&57.73&59.87&58.40&\textbf{60.53}\\
  \hline
  \multirow{1}*{Aging Quality ($\uparrow$)} &\textbf{Overall}&36.80&44.15&40.18&\textbf{55.54}\\
  \bottomrule
  \end{tabular}
\end{table}

\textbf{Human Evaluation.} To provide more reliable quantitative analysis, we perform a human evaluation to compare different methods. Following SAM~\cite{alaluf2021only}, we compare different methods on three age groups: \textit{`a toddler'}, \textit{`an adult'}, and \textit{`the elderly'}. Specifically, we invite 60 volunteers and ask them to score the results of different methods, according to aging accuracy, identity preservation and aging quality. 
Each volunteer is randomly allocated 30 reference image sets. Note that all images are randomly selected. We provide the human evaluation results in Table~\ref{tab:user}. Overall represents the results of all three age groups. As expected, our PADA consistently performs best in all three metrics. We observe that the aging accuracy of DLFS\cite{he2021disentangled} on \textit{`a toddler'} is a little better than ours, while the identity preservation of SAM\cite{alaluf2021only} on \textit{`a toddler'} is a little better than ours. This is because the aging accuracy and identity preservation may conflict when we translate an adult into a toddler. 

\begin{figure}
\begin{center}
\includegraphics[width=1\linewidth]{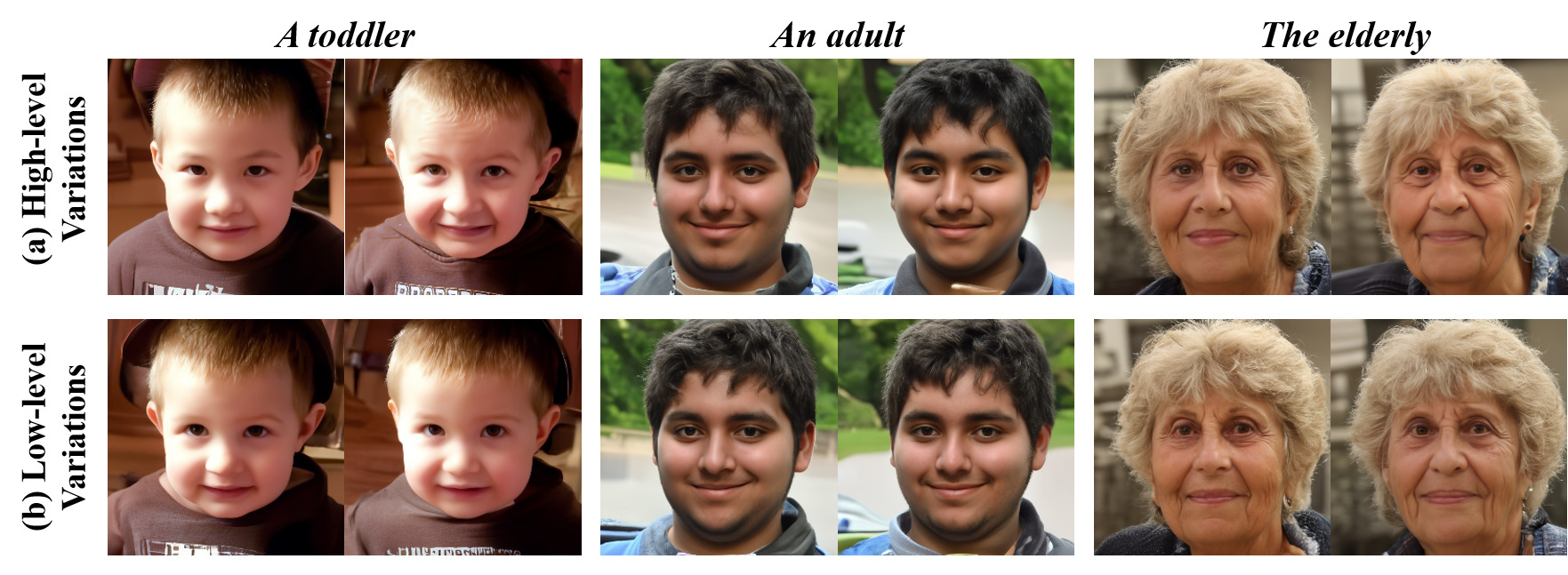}
\end{center}
   \caption{Pluralistic face aging results of PADA. The first row shows the diverse aging results with high-level variations. The second row shows the results with low-level variations.}
\label{fig:diverse}
\end{figure}

\begin{figure}
\begin{center}
\includegraphics[width=1\linewidth]{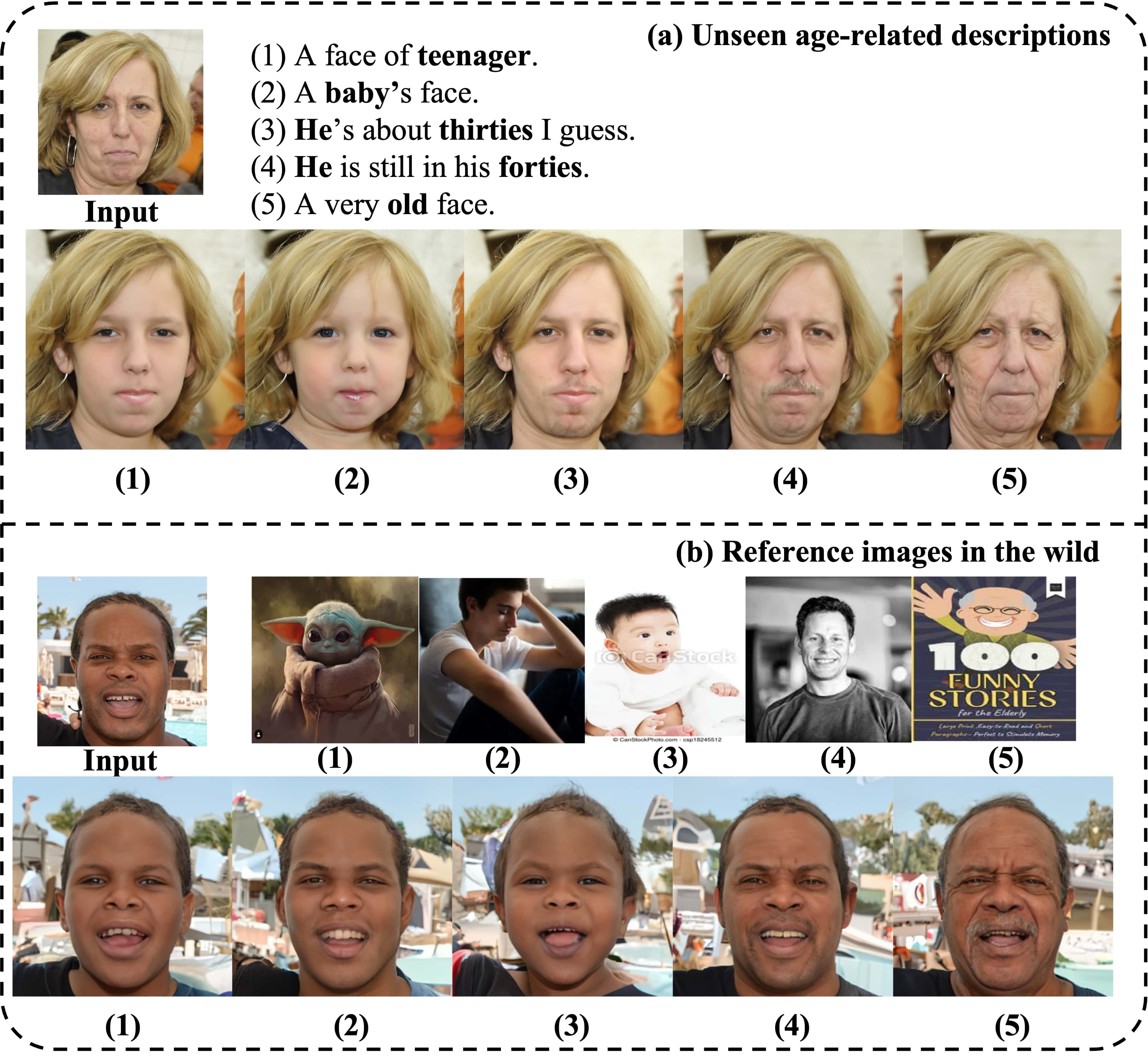}
\end{center}
   \caption{Face aging conditioned on unseen age-related descriptions and reference images in the wild. (a) Despite never being trained with texts of `a very old face', our PADA still yields plausible face aging results. (b) We can utilize arbitrary reference images to guide the aging process.}
\label{fig:opentext}
\end{figure}

\subsection{Ablation Study}

\textbf{Diversity Exploration.} Due to the sequential denoising reverse process and the probabilistic aging embedding learning, our PADA generates pluralistic aging results with both low-level variations and high-level variations. We show the diverse aging results in Fig.~\ref{fig:diverse}. The high-level variations are relevant to face shape and skin color, while the low-level variations are relevant to the types and locations of wrinkles.

\textbf{Text-guided and Reference-guided Face Aging.} Thanks to our strategy of probabilistic aging embedding learning in CLIP latent space, our PADA can achieve face aging conditioned on the unseen age-related text descriptions or arbitrary face images from the Internet. Thus, both the text-driven and reference-driven interfaces for face aging are provided. The results based on the open-world age descriptions or arbitrary unseen facial images are shown in Fig.~\ref{fig:opentext}. Although our PADA has not seen both these two variants during training, it still can generate plausible face aging results.

\begin{figure}
\begin{center}
\includegraphics[width=1\linewidth]{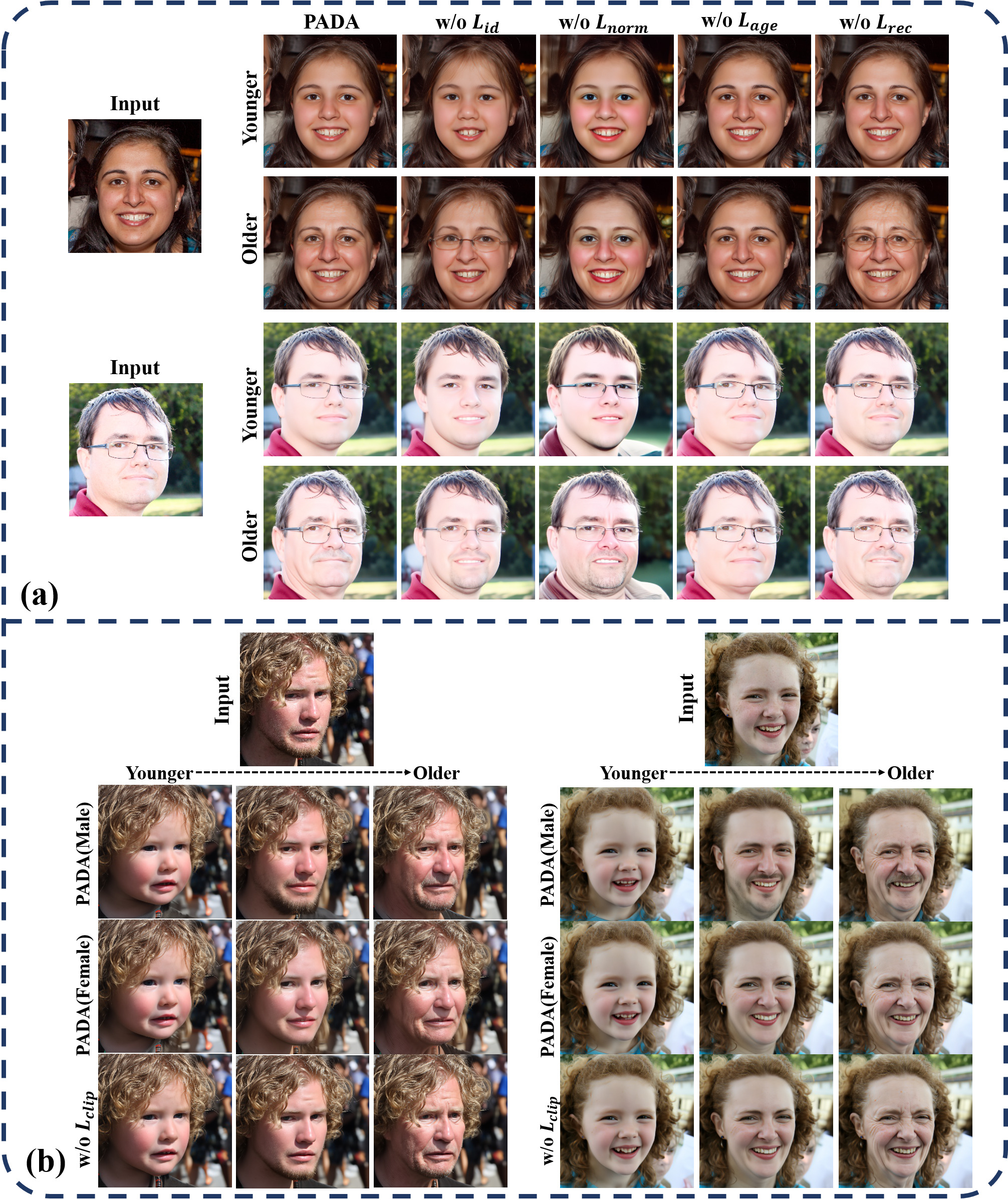}
\end{center}
   \caption{Visual comparisons of our method with its variants.}
\label{fig:ablation}
\end{figure}

\textbf{The Effectiveness of Losses.}
We report the qualitative visualization results in Fig.~\ref{fig:ablation} for a comprehensive comparison between our PADA and its five variants. The quantitative comparison is included in supplementary materials.
As shown in Fig.~\ref{fig:ablation} (a), 
without $L_{id}$ loss, much identity information is lost during aging process. The pose and expression information are also dramatically destroyed. 
Without $L_{norm}$ loss, the generation quality is degraded and unrealistic images with artifacts are produced, e.g., the blush on the face.
The lack of $L_{age}$ leads to almost negligible aging changes. 
$L_{rec}$ is randomly applied for each batch during training, but it can preserve age-irrelevant information.

As shown in Fig.~\ref{fig:ablation} (b),
$L_{clip}$ improves the aging fidelity from two perspectives. (I) It makes our PADA learn more aging details from the pre-trained CLIP, e.g., the beards and necklines. Additionally, it leads to better shape transformation for babies; 
(II) It can accurately model aging process for different genders. Without $L_{clip}$, our PADA can hardly model the clearly gender-specific aging process.
These indicate that each component of PADA is essential.

Table~\ref{tab:ab_quan} further presents the performance on aging accuracy and identity preservation of different variants of our PADA on FFHQ-AT test set.
As expected, without the preservation loss: $L_{id}$, $L_{norm}$, or $L_{rec}$, the performance of identity preservation decreases. Without the age fidelity loss: $L_{age}$ or $L_{clip}$, the performance of aging accuracy has dropped. These indicate that each component in our method is essential for synthesizing photo-realistic aging results. 

Additionally, we observe that without the age fidelity loss: $L_{age}$ or $L_{clip}$, the model can achieve better identity preservation. Meanwhile, without the preservation loss: $L_{id}$ or $L_{norm}$, the performance of aging accuracy is improved. All of the above phenomena are reasonable. This is because that there may be conflict between aging accuracy and identity preservation. Specifically, according to the survey of age-invariant face recognition~\cite{park2010age}, both the shape and texture changes degrade the performance of face recognition systems. The employ of age fidelity loss leads to the shape and texture changes, which degrades the performance of identity preservation. While the employ of the preservation loss inhibits the shape or texture changes, which degrades the performance of aging accuracy.
Our PADA better balances the aging accuracy and identity preservation, achieving plausible face aging with pretty identity preservation.

\begin{table}\scriptsize
\centering
\caption{Model comparisons on FFHQ-AT.}
\label{tab:ab_quan}
\renewcommand{\arraystretch}{1} 
 \setlength\tabcolsep{2.5pt}
\begin{tabular}{c|c|c}
  \toprule
  Method & Age MAE ($\downarrow$) & ID Preservation ($\uparrow$) \\
  \hline
  \multirow{1}*{w/o $L_{id}$}& \textbf{8.70} & 0.4766 \\ 
   \multirow{1}*{w/o $L_{norm}$}& 9.23 & 0.6446 \\
  \multirow{1}*{w/o $L_{rec}$}& 9.28 & 0.4516 \\
  \multirow{1}*{w/o $L_{clip}$} & 11.07 & 0.7046  \\
  \multirow{1}*{w/o $L_{age}$} & 16.62 & \textbf{0.7176} \\
  \multirow{1}*{w/o $L_{kl}$} & 9.99 & 0.6736  \\
  \hline
  \multirow{1}*{Ours} & 9.19 & 0.6516   \\
  \bottomrule
  \end{tabular}
\end{table}


\section{Conclusions}

In this paper, we have proposed a CLIP-driven Pluralistic Aging Diffusion Autoencoder (PADA) to achieve diverse plausible face aging with both low-level stochastic variations and high-level aging semantic variations. To produce stochastic low-level aging details, we resort to diffusion models by sequential denoising reverse. To learn diverse high-level aging patterns, we present Probabilistic Aging Embedding (PAE) in the common CLIP latent space, where a text-guided KL-divergence loss is imposed to guide the learning of distribution. Our PADA can achieve pluralistic face aging conditioned on both the open-world age-related descriptions and arbitrary unseen facial images. Extensive experiments demonstrate that our method obtains state-of-the-art face aging results in terms of generation quality and diversity.

\paragraph{Acknowledgment} This work is sponsored by Beijing Nova Program (Z211100002121106), National Natural Science Foundation of China (Grant No.62006228, Grant No. 62176025), Youth Innovation Promotion Association CAS (Grant No.2022132), and CAAI-Huawei MindSpore Open Fund.

\clearpage
{\small
\bibliographystyle{ieee_fullname}
\bibliography{egbib}
}
\clearpage

\section*{Appendix}

\appendix

In this appendix, we first introduce the theory validation in Sec.~\ref{sec:theory}. Then, we show the training and inference algorithms in Sec.~\ref{sec:methods}. Additional qualitative and quantitative comparison results are shown in Sec.~\ref{sec:comp}. In this section, we propose a straightforward technique for achieving gender transformation utilizing the proposed PAEs. Meanwhile, we conduct disentanglement experiments in terms of different timesteps and compare PADA with previous method on Morph and CACD2000. Also, we measure the diversity boundary of PADA. Finally, we show more pluralistic face aging results in Sec.~\ref{sec:aging}, including reference-guided face aging, text-guided face aging, diverse face aging, and intermediate generation results of diffusion decoder.

\section{Theory Validation}\label{sec:theory}
\textbf{Theorem 1.} In the normalized CLIP latent space, according to the Law of Cosines, the Euclidean distance $D(e^{age}, e^{txt})$ between probabilistic aging embedding $e^{age}$ and text-based age representation $e^{txt}$ is optimally equivalent to the cosine similarity.

\textbf{Proof.} In practice, we normalize all the features in the CLIP latent space by $L_2$ norm. Hence, according to the Law of Cosines, the equivalent form of $D(e^{age}, e^{txt})$ can be rewritten as:
\begin{equation}
\begin{aligned}
    D&(e^{age}, e^{txt}) = \left\| e^{age} - e^{txt} \right\|^2_2 
    \\=  &\left\| e^{age} \right\|^2_2 + \left\| e^{txt} \right\|^2_2 - 2 \left\| e^{age} \right\|_2 \left\| e^{txt} \right\|_2 cos(e^{age}, e^{txt})
    \\= & 2-2cos(e^{age}, e^{txt}) \nonumber 
\end{aligned}
\end{equation}
Therefore, when calculating the loss $L_{tKL}$, the optimization objectives for the Euclidean distance $D(e^{age}, e^{txt})$ and cosine distance  $-cos(e^{age}, e^{txt})$ are equivalent.

\textbf{Theorem 2.} For directly sampling PAE from text-based age prior, the Euclidean distance $D$ between the probabilistic aging embedding $e^{age}$ and corresponding aging text representation $e^{txt}$ for $\forall{m^*}$ satisfies $D(e^{txt}, e^{age})\leq m^*$ with probability at least:
\begin{equation}
\begin{aligned}
    &Prob ( D(e^{txt}, e^{age})\leq m^* ) \\ 
    &= 1 - \int_{-1}^{1 - \frac{m^*}{2}  - \frac{m^*}{2\epsilon}} \frac{\Gamma(d/2+1/2)}{\sqrt{\pi}\Gamma(d/2)}(1-x^2)^{d/2-1}dx,\nonumber 
\end{aligned}
\end{equation}
where $\Gamma(\cdot)$ is Gamma function, i.e.\ $\Gamma(\cdot)=\int_{0}^{\infty}x^{t-1}e^{-x}dx$. $d$ is the dimension of input feature, $\epsilon$ is hyperparameter for sampling intensity, and $\eta$ is normalized sampling from normal Gaussian distribution.

\textbf{Proof.} In practice, we normalize the features in CLIP latent space by $L_2$ norm. 

According to the Law of Cosines, we get:
\begin{equation}
\begin{aligned}
    D(e^{txt}, e^{age}) &=\left\| e^{txt} - e^{age} \right\|_2^2\\
    &= 2\left(1-\frac{(e^{txt})^Te^{age}}{ \left\| e^{txt} \right\|_2 \left\| e^{age} \right\|_2} \right ) \\
    &= 2\left(1-\frac{\left\| e^{txt} \right\|^2_2 + \epsilon \cdot (e^{txt})^T \eta}{ \left\| e^{txt} + \epsilon   \cdot \eta \right\|_2}\right ) \\
    &= 2\left(1- \frac{1 + \epsilon \cdot (e^{txt})^T\eta}{ \left\| e^{txt} + \epsilon  \cdot {\eta} \right\|_2}\right ) \\
    &\leq 2\left(1- \frac{1 + \epsilon \cdot (e^{txt})^T {\eta}}{ \left\| e^{txt} \right\|_2 + \epsilon \left\| \eta \right\|_2}\right ) \\
    &= 2\left(1- \frac{1 + \epsilon \cdot (e^{txt})^T {\eta}}{ 1 + \epsilon} \right) \nonumber
\end{aligned}
\end{equation}

Therefore, we find an lower bound for our original probability:
\begin{equation}
\begin{aligned}
    &Prob ( D(e^{txt}, e^{age})\leq m^* ) \\ 
    &\geq  Prob \left( 2\left(1- \frac{1 + \epsilon \cdot (e^{txt})^T {\eta}}{ 1 + \epsilon} \right)· \leq m^* \right )\\
    &= 1 - Prob \left( (e^{txt})^T \eta  \leq  1 - \frac{m^*}{2}  - \frac{m^*}{2\epsilon}\right ) \nonumber
\end{aligned}
\end{equation}

In \cite{cho2009inner}, the Cumulative Distribution Function (CDF) of the inner product of two random vectors, i.e.\ $x=u^Tv$ on a standard unit sphere is:

\begin{equation}
\begin{aligned}
    F(x)=\int_{-1}^{x} \frac{\Gamma(d/2+1/2)}{\sqrt{\pi}\Gamma(d/2)}(1-x^2)^{d/2-1}dx
\end{aligned}
\end{equation}

Thus, we complete our proof:
\begin{equation}
\begin{aligned}
    &Prob \left( D(e^{txt}, e^{age})\leq m^* \right ) \\ 
    &\geq 1 - Prob \left( (e^{txt})^T \eta \leq 1 - \frac{m^*}{2}  - \frac{m^*}{2\epsilon}\right ) \\
    &= 1 - \int_{-1}^{1 - \frac{m^*}{2}  - \frac{m^*}{2\epsilon}} \frac{\Gamma(d/2+1/2)}{\sqrt{\pi}\Gamma(d/2)}(1-x^2)^{d/2-1}dx\nonumber 
\end{aligned}
\end{equation}

\section{Details on Methods}\label{sec:methods}
For detailed explanation, we show the Training pipeline and Inference pipeline of our PADA in Algorithm \ref{alg1} and \ref{alg2}, respectively. 
\begin{algorithm}[h]
	\caption{Training stage of PADA: given a pre-trained conditional noise prediction network $\epsilon(x_t,t,z)$, a pre-trained semantic encoder $E_{sem}$, and a pre-trained CLIP image/text encoder $E_{img}$/$E_{txt}$}
    \renewcommand{\algorithmicrequire}{\textbf{Input:}}
	\renewcommand{\algorithmicensure}{\textbf{Output:}}
	\label{alg1}
	\begin{algorithmic}[1]
        \REQUIRE source image $x_0^{src}$, reference image $x_0^{ref}$, reference text $t^{ref}$, diffusion step $T$
        \ENSURE $\theta^*$ (the parameters of CLIP-guided Age Encoder $E_{age}$)
        \REPEAT
            \STATE $t\sim Uniform({1,...,T})$
            \STATE $x^{src}_t\sim \mathcal{N}(\sqrt{\overline{\alpha}_t}x_0^{src}, (1-\overline{\alpha}_t)\bm{I})$
            \STATE $x^{ref}_t\sim \mathcal{N}(\sqrt{\overline{\alpha}_t}x_0^{ref}, (1-\overline{\alpha}_t)\bm{I})$
            \STATE $z^{src},z^{ref} \leftarrow E_{sem}(x_0^{src}),E_{sem}(x_0^{ref})$
            \STATE $\hat{x}^{src}_0 \leftarrow \frac{x^{src}_t}{\sqrt{\alpha}_t} - \frac{\sqrt{1-{\alpha}_t}}{\sqrt{\alpha}_t} \epsilon(x^{src}_t, t, z^{src})$
            \STATE $\hat{x}^{ref}_0 \leftarrow \frac{x^{ref}_t}{\sqrt{\alpha}_t} - \frac{\sqrt{1-{\alpha}_t}}{\sqrt{\alpha}_t} \epsilon(x^{ref}_t, t, z^{ref})$
            \STATE $r\leftarrow random(0,1)$
            \IF{$r \leq 0.5$}
                \STATE $z^{age} \leftarrow E_{age}\left ( E_{img}(x_0^{ref})\right )$
            \ELSIF{$r \leq 0.8$}
                \STATE $z^{age} \leftarrow E_{age}\left ( E_{txt}(t^{ref})\right )$
            \ELSE
                \STATE $z^{age} \leftarrow E_{age}\left ( E_{img}(x_0^{src})\right )$
            \ENDIF
            \STATE $z^{tar} \leftarrow z^{src} + z^{age}$
            \STATE $\hat{x}^{tar}_0 \leftarrow \frac{x^{src}_t}{\sqrt{\alpha}_t} - \frac{\sqrt{1-{\alpha}_t}}{\sqrt{\alpha}_t} \epsilon(x^{src}_t, t, z^{tar})$
            \STATE Compute total loss $L(\hat{x}^{tar}_0, \hat{x}^{src}_0,\hat{x}^{ref}_0, t^{ref})$
            \STATE Take a gradient step on $\nabla_{\theta}L$
        \UNTIL coveraged
	\end{algorithmic}  
\end{algorithm}

\begin{algorithm}[h]
	\caption{Inference stage of PADA: given a pre-trained conditional noise prediction network $\epsilon(x_t,t,z)$, a semantic encoder $E_{sem}$, a pre-trained CLIP image/text encoder $E_{img}$/$E_{txt}$, and lerned CLIP-guided age encoder $E_{age}$}
    \renewcommand{\algorithmicrequire}{\textbf{Input:}}
	\renewcommand{\algorithmicensure}{\textbf{Output:}}
	\label{alg2}
	\begin{algorithmic}[1]
        \REQUIRE source image $x_0^{src}$, reference image $x^{ref}$ or reference text $t^{ref}$, generation step $T$
		\STATE $x^{tar}_T\sim \mathcal{N}(\bm{0}, \bm{I})$
            \STATE $z^{src} \leftarrow E_{sem}(x_0^{src})$
            \IF{$image-guided$}
                \STATE $z^{age} \leftarrow E_{age}\left ( E_{img}(x_0^{ref})\right)$
            \ELSIF{$text-guided$}
                \STATE $z^{age} \leftarrow E_{age}\left ( E_{txt}(t^{ref})\right )$
            \ENDIF
            \STATE $z^{tar} \leftarrow z^{src} + z^{age}$
            \FOR{ $t=T,...,1$}
                \STATE $\hat{x}_0^{tar} \leftarrow \frac{x^{tar}_t}{\sqrt{\alpha}_t} - \frac{\sqrt{1-{\alpha}_t}}{\sqrt{\alpha}_t} \epsilon(x^{tar}_t, t, z^{tar})$
                \STATE $x^{tar}_{t-1} \leftarrow \sqrt{\alpha_{t-1}}\hat{x}_0^{tar} + \sqrt{1-\alpha_{t-1}}\cdot \epsilon(x^{tar}_t, t, z^{tar})$ 
            \ENDFOR
            \ENSURE target aging result $x_0^{tar}$.
	\end{algorithmic}  
\end{algorithm}

\begin{figure*}
\begin{center}
\includegraphics[width=1\linewidth]{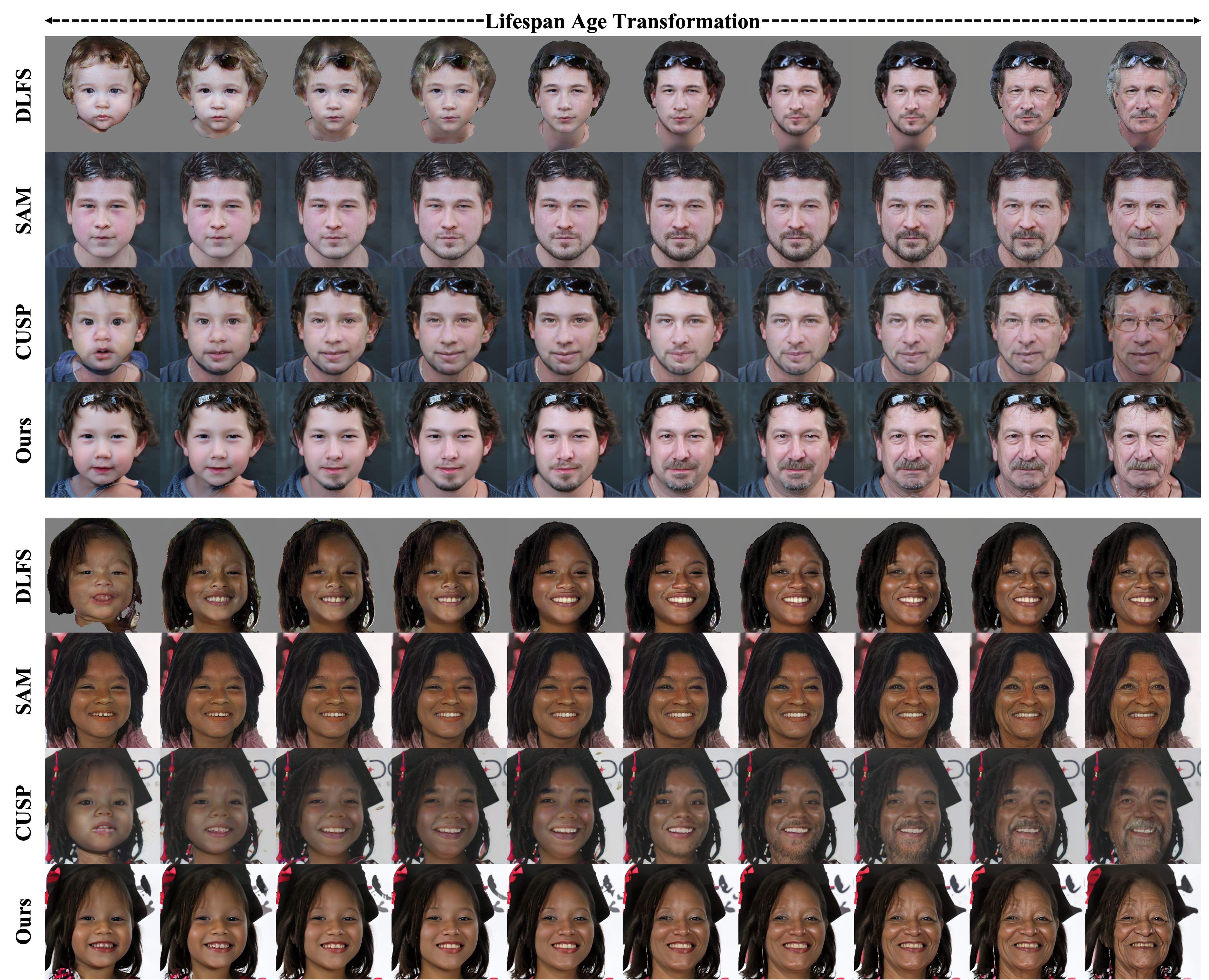}
\end{center}
   \caption{Continuous face aging by interpolation in latent space. Best viewed zoomed-in.}
\label{fig:interpolation}
\end{figure*}

\begin{figure*}
\begin{center}
\includegraphics[width=1\linewidth]{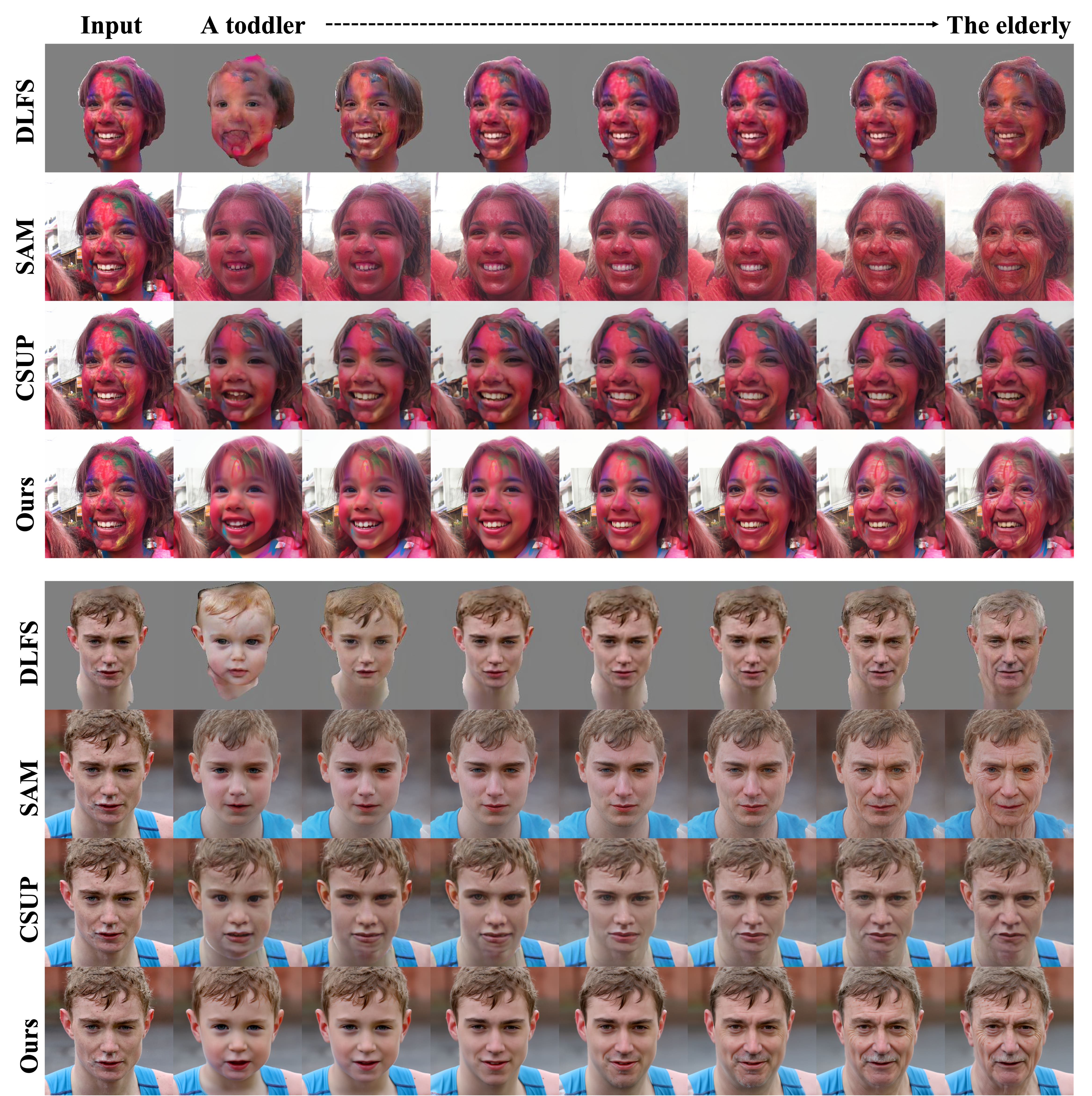}
\end{center}
   \caption{More comparison results with DLFS~\cite{he2021disentangled}, SAM~\cite{alaluf2021only}, and CUSP~\cite{gomez2022custom} on FFHQ-AT test set.}
\label{fig:1}
\end{figure*}

\begin{figure*}
\begin{center}
\includegraphics[width=1\linewidth]{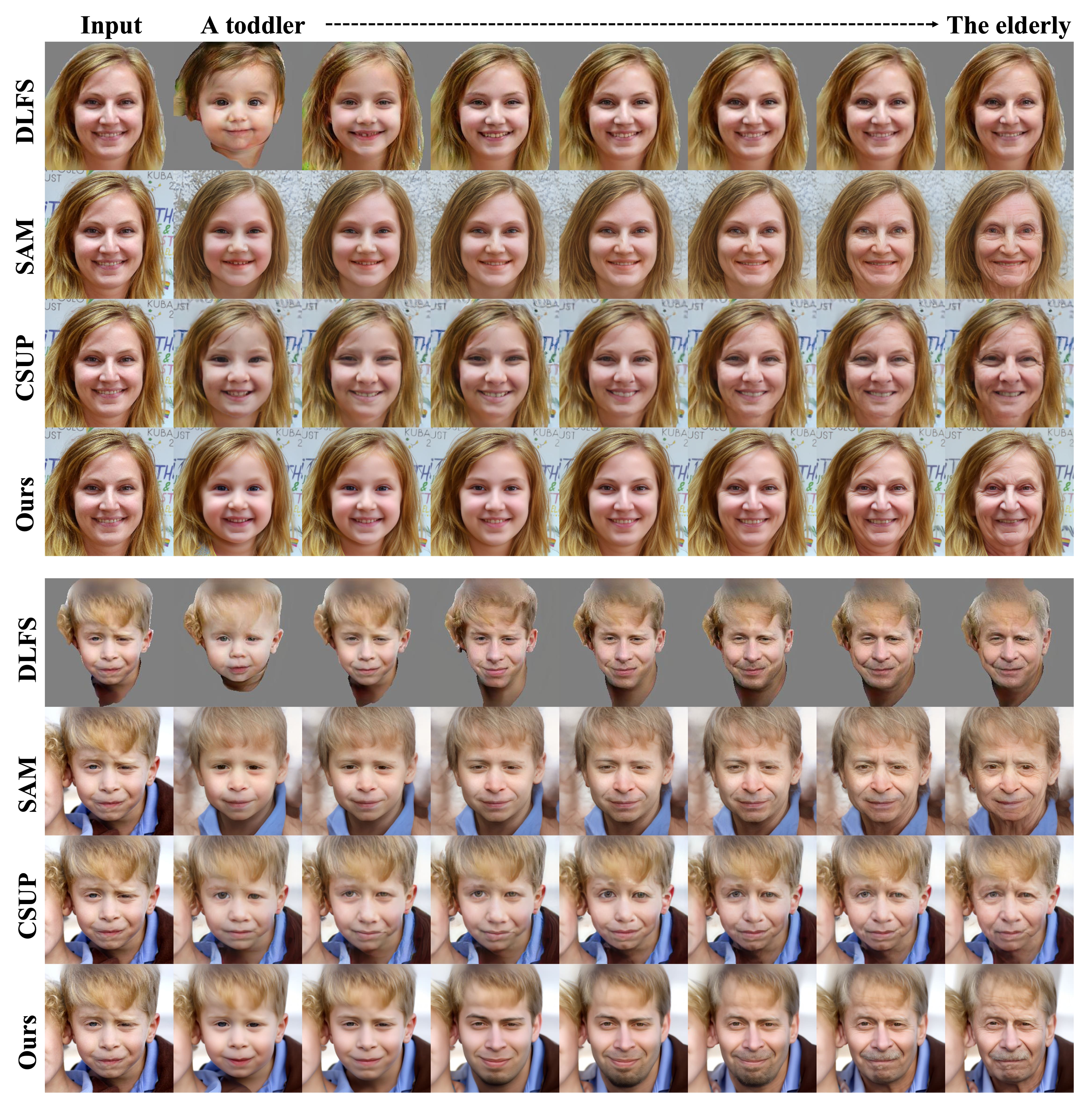}
\end{center}
   \caption{More comparison results with DLFS~\cite{he2021disentangled}, SAM~\cite{alaluf2021only}, and CUSP~\cite{gomez2022custom} on FFHQ-AT test set.}
\label{fig:2}
\end{figure*}

\begin{figure*}
 \begin{center}
 \includegraphics[width=1\linewidth]{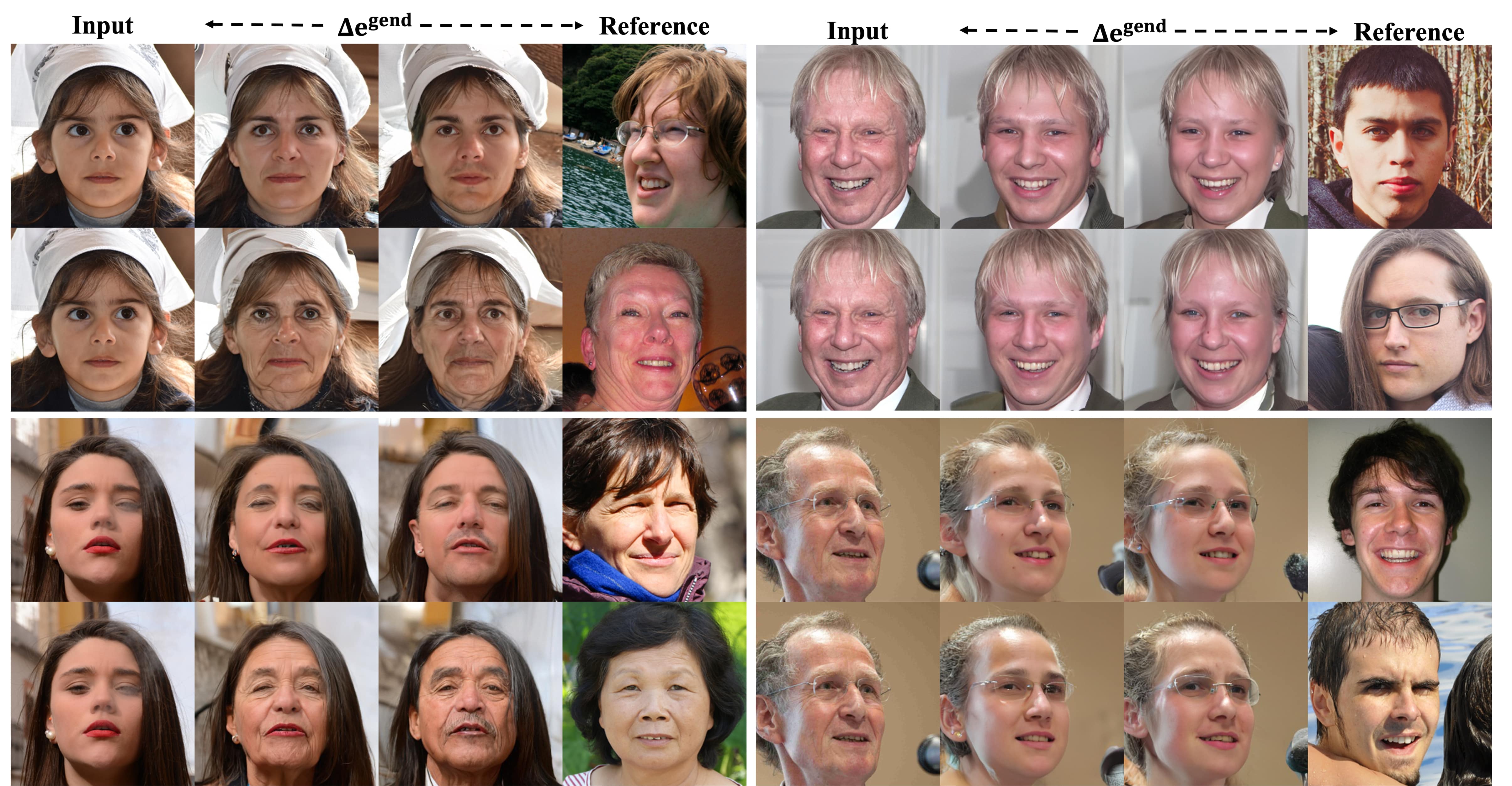}
 \end{center}
    \caption{Results with/without gender adjustment.}
\label{fig:gender_transformation}
 \end{figure*}

\begin{figure*}
\begin{center}
\includegraphics[width=1\linewidth]{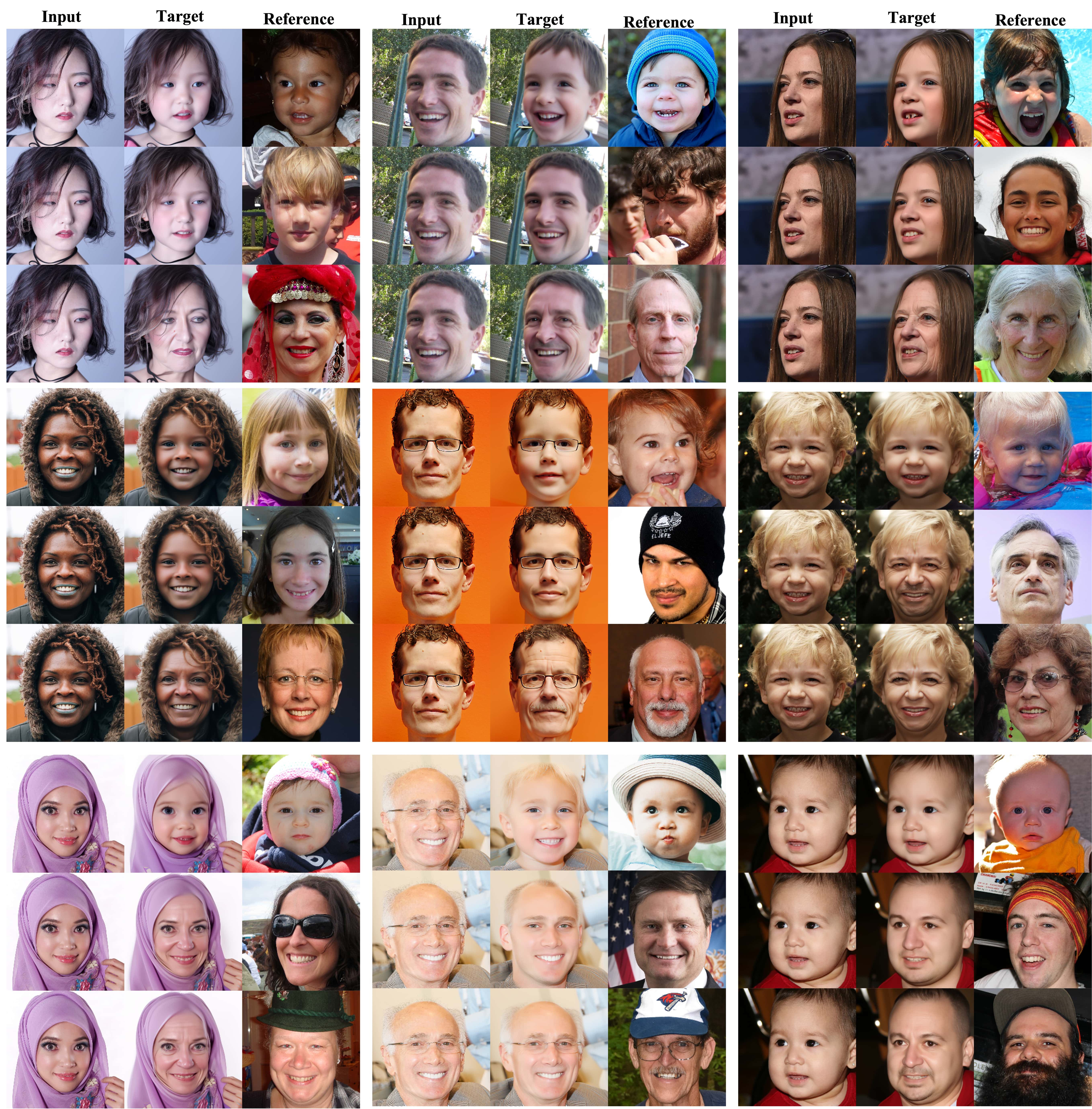}
\end{center}
   \caption{Reference-guided aging results on FFHQ-AT test set.}
\label{fig:3}
\end{figure*}

\begin{figure*}
\begin{center}
\includegraphics[width=1\linewidth]{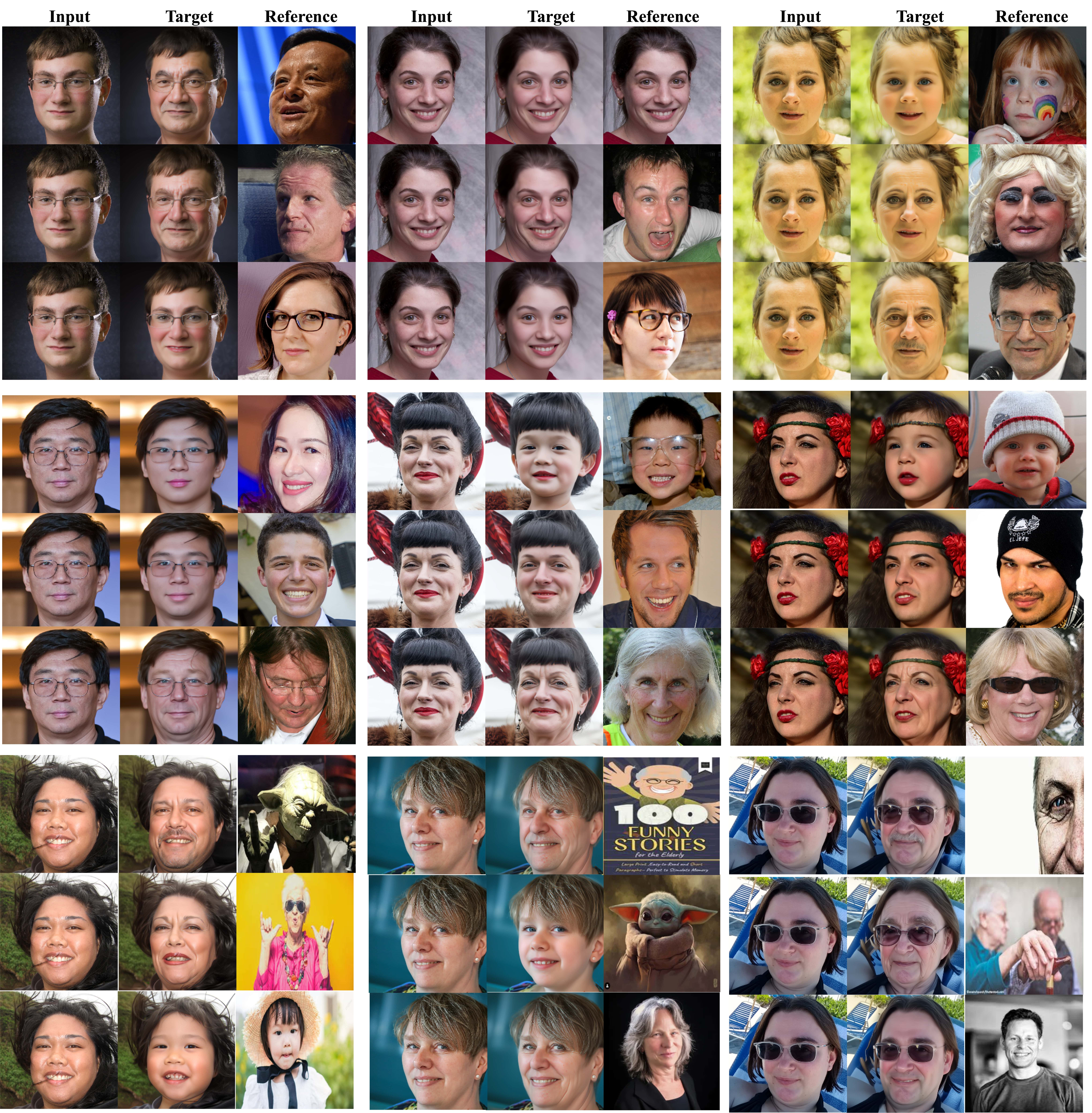}
\end{center}
   \caption{Reference-guided aging results on FFHQ-AT test set.}
\label{fig:4}
\end{figure*}

\section{Qualitative and Quantitative Comparisons}\label{sec:comp}

\paragraph{}We compare the continuous face aging capabilities of our PADA with DLFS~\cite{he2021disentangled}, SAM~\cite{alaluf2021only}, and CUSP~\cite{gomez2022custom} on CelebA-HQ test set in Fig.~\ref{fig:interpolation}. Obviously, both the aging accuracy and age-irrelevant information preservation of our method are superior to these methods. Meanwhile, in Fig.~\ref{fig:1} and Fig.~\ref{fig:2}, we show more comparison results with the three state-of-the-art methods on FFHQ-AT test set.

\paragraph{Gender Adjustment.} As our PAE is proposed in CLIP latent space and incorporates gender information during aging training, we are able to perform gender adjustment using the formula $e^{rec}=e^{age} \pm \Delta e^{gend}$, where $\Delta e^{gend}=e^{m}-e^{w}$ and $e^{m}$ and $e^{w}$ correspond to the embeddings of `man's face' and `woman's face', respectively. Fig.~\ref{fig:gender_transformation} displays the results obtained after applying gender adjustment. More results can be found in Fig.~\ref{fig:3} and Fig.~\ref{fig:4}.

\begin{table}[h]
\centering
\caption{Quantitative analysis of diversity boundaries.}
\label{tab:bd}
\renewcommand{\arraystretch}{1} 
 \setlength\tabcolsep{2.5pt}
\begin{tabular}{l|c|cccc}
  \toprule
  \multirow{2}{*}{Variance} & \multirow{2}{*}{+Low-level} & \multicolumn{4}{c}{+Low-level+High-level ($\epsilon$)}\\\cline{3-6}
  & & 0.01 & 0.1 & 0.25 & 0.5 \\
  \hline
  LPIPS ($\uparrow$) & 0.189  & 0.193 & 0.194 & 0.199 & 0.203 \\ 
  ID ($\uparrow$) & 0.668 & 0.649 & 0.633 & 0.617 & 0.593 \\
  \bottomrule
  \end{tabular}
\end{table}

\paragraph{Diversity Boundary.} Following PICNet\cite{zheng2019pluralistic}, we evaluate our diversity with LPIPS. The average score is calculated between 1k pairs generated with and without variations. In Table~\ref{tab:bd}, as the sampling intensity $\epsilon$ of high-level variations increases, the diversity score increases, while the ID score slightly decreases. These indicate the promising performance of our PADA for generating diverse results while preserving identity.

\begin{figure*}[h]
\centering
\includegraphics[width=1\linewidth]{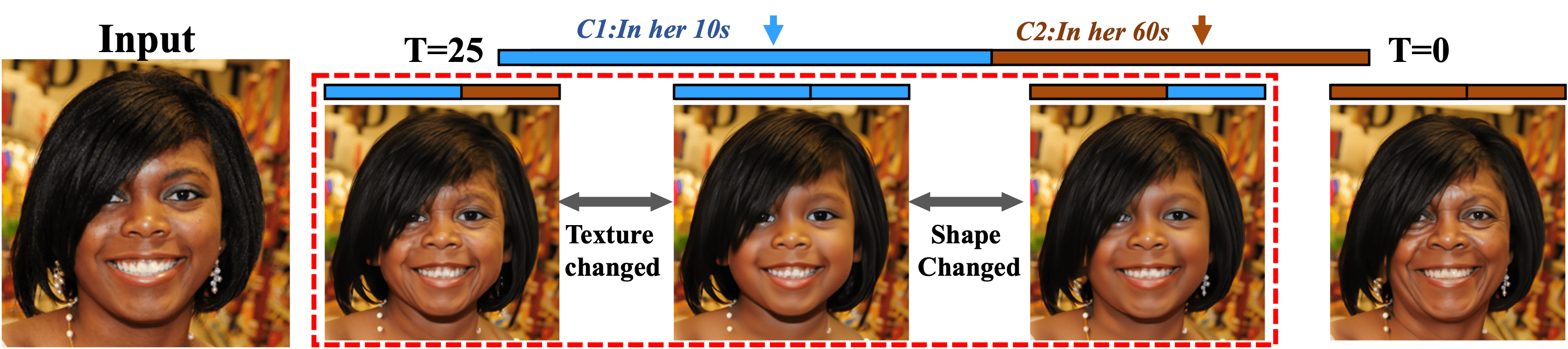}
   \caption{Manipulation at different time $(T)$. }
\label{fig:effect}
\end{figure*}

\paragraph{Disentanglement in PADA.}
As shown in Fig~\ref{fig:effect}, the early denoising steps (T=25 to T=10) prioritize shape, while the later steps (T=10 to T=0) prioritize texture. 
For example, if we replace C1 with C2 in later denoising steps, the generated texture corresponds to C2, while the generated shape corresponds to C1. This verifies the effectiveness of PADA for face aging.

\begin{figure*}[h]
\centering
\includegraphics[width=1\linewidth]{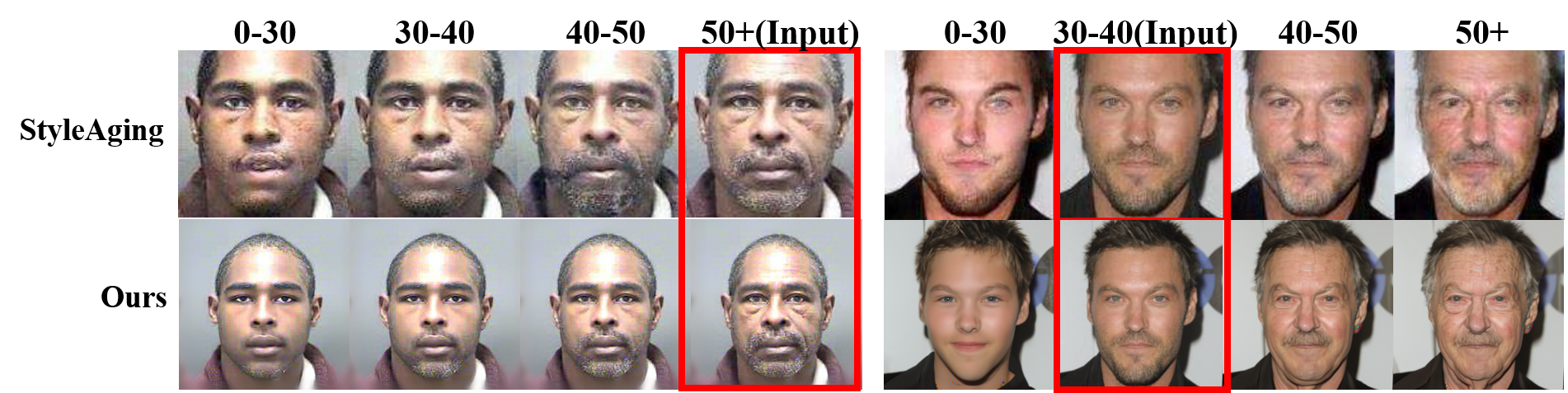}
   \caption{Comparisons on Morph(left) and CACD2000(right). }
\label{fig:morph}
\end{figure*}

\paragraph{Comparison with StyleAging~\cite{georgopoulos2020enhancing} on Morph and CACD2000.} We compare PADA with StyleAging~\cite{georgopoulos2020enhancing}. Since there is the domain bias between Morph and FFHQ, so we first finetune the pretrained DiffAE on Morph dataset with 2 epochs. Compared with StyleAging~\cite{georgopoulos2020enhancing}, our method achieves better generation quality and aging fidelity. The results are shown in Fig~\ref{fig:morph}.

\begin{figure*}[h]
\centering
\includegraphics[width=1\linewidth]{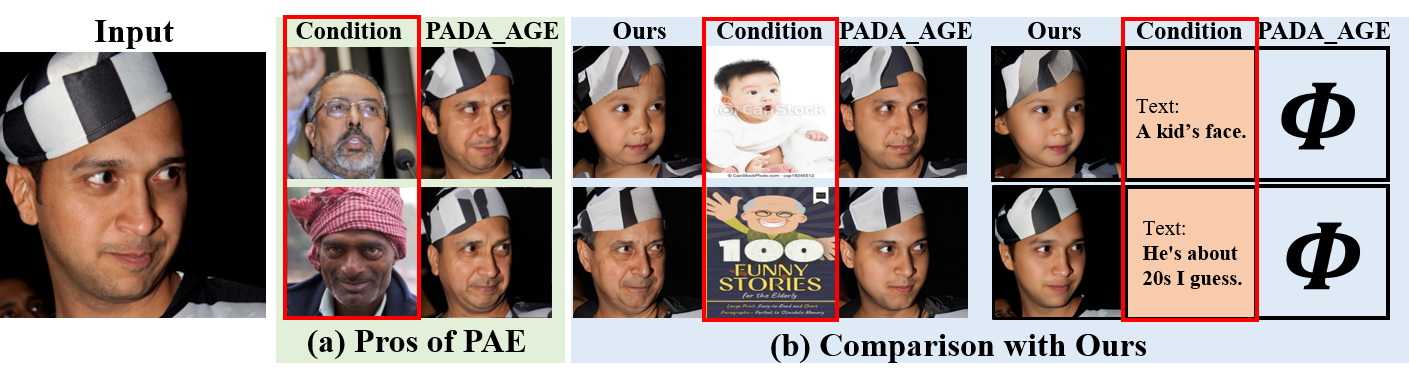}
   \caption{Compared with other feature space. }
\label{fig:agespace}
\end{figure*}

\paragraph{Effectiveness of CLIP Space.} To validate the effectiveness of CLIP feature space, we replace the CLIP image encoder with a pre-trained age estimator and adopt PAE in its latent space(called PADA\_AGE). As shown in Fig.~\ref{fig:agespace}, it can generate diverse aging results, indicating the effectiveness of our PAE. However, PADA\_AGE has limited flexibility, as it cannot directly generate images conditioned on exact age. 
Additionally, its generalization ability is limited, as it fails at face aging conditioned on reference images in the wild.

\section {Pluralistic Face Aging}\label{sec:aging}

\begin{figure*}
\begin{center}
\includegraphics[width=1\linewidth]{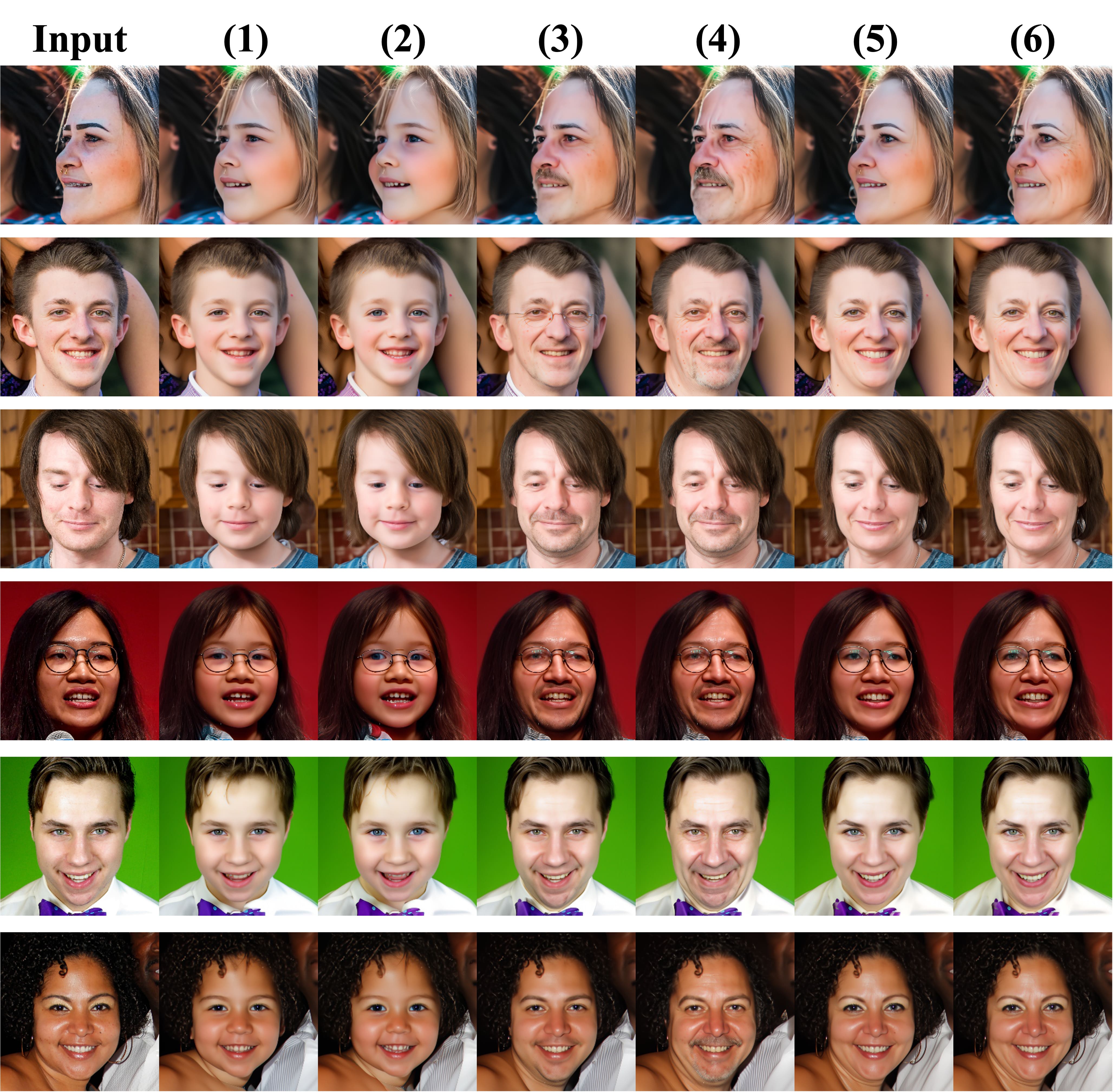}
\end{center}
   \caption{Text-guided aging results on FFHQ-AT test set. We apply different unseen age-related text descriptions as conditions. Concretely, (1) "a quite young boy", (2) "a daughter aged five", (3) "a face in his early forties ", (4) "a face in his late forties", (5) "a face in her early forties", (6) "a face in her late forties". }
\label{fig:text}
\end{figure*}

\begin{figure*}
\begin{center}
\includegraphics[width=1\linewidth]{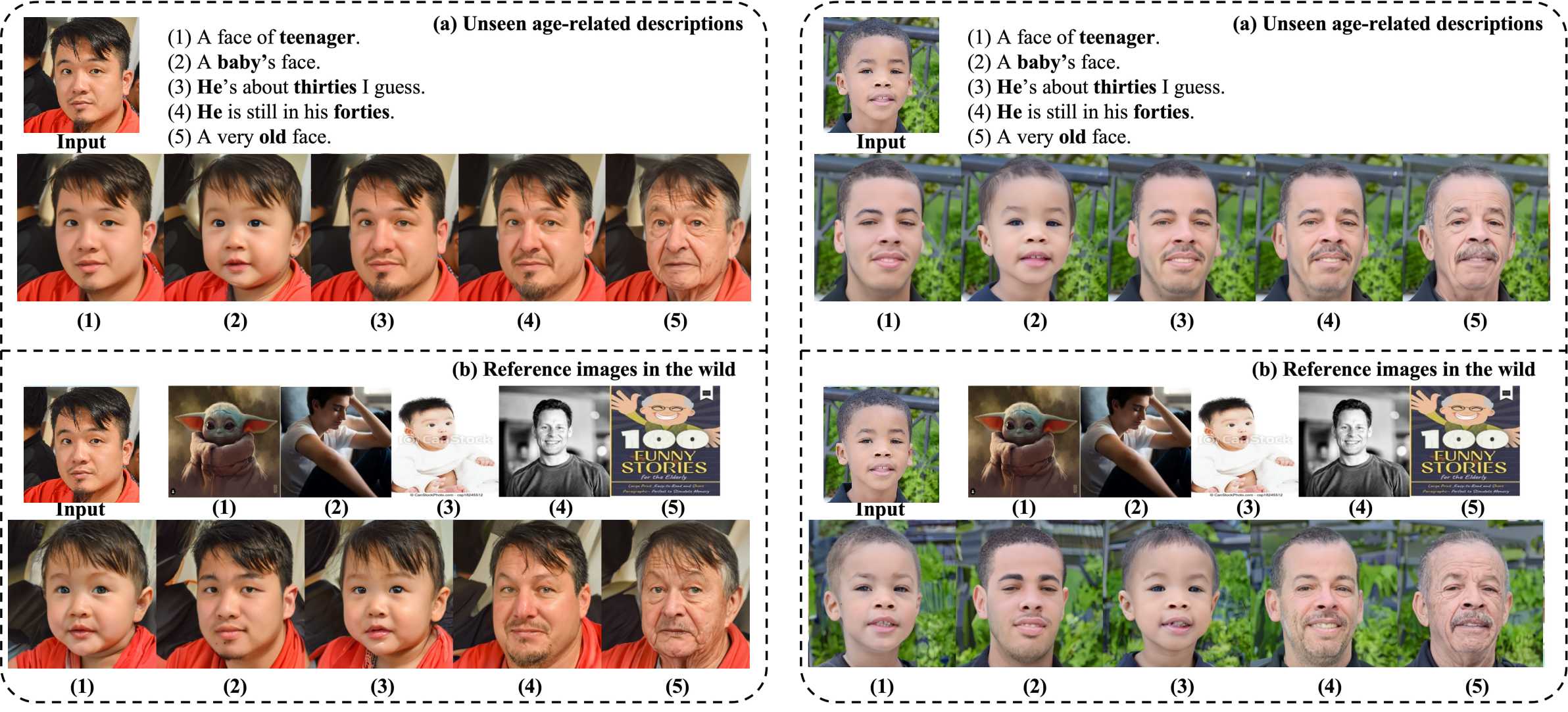}
\end{center}
   \caption{Face aging conditioned on unseen age-related descriptions and reference images in the wild. (a) Despite never being trained with texts of `a very old face', our PADA still yields plausible face aging results. (b) We can utilize arbitrary reference images to guide the aging process. }
\label{fig:in_the_wild}
\end{figure*}

\begin{figure*}
\begin{center}
\includegraphics[width=1\linewidth]{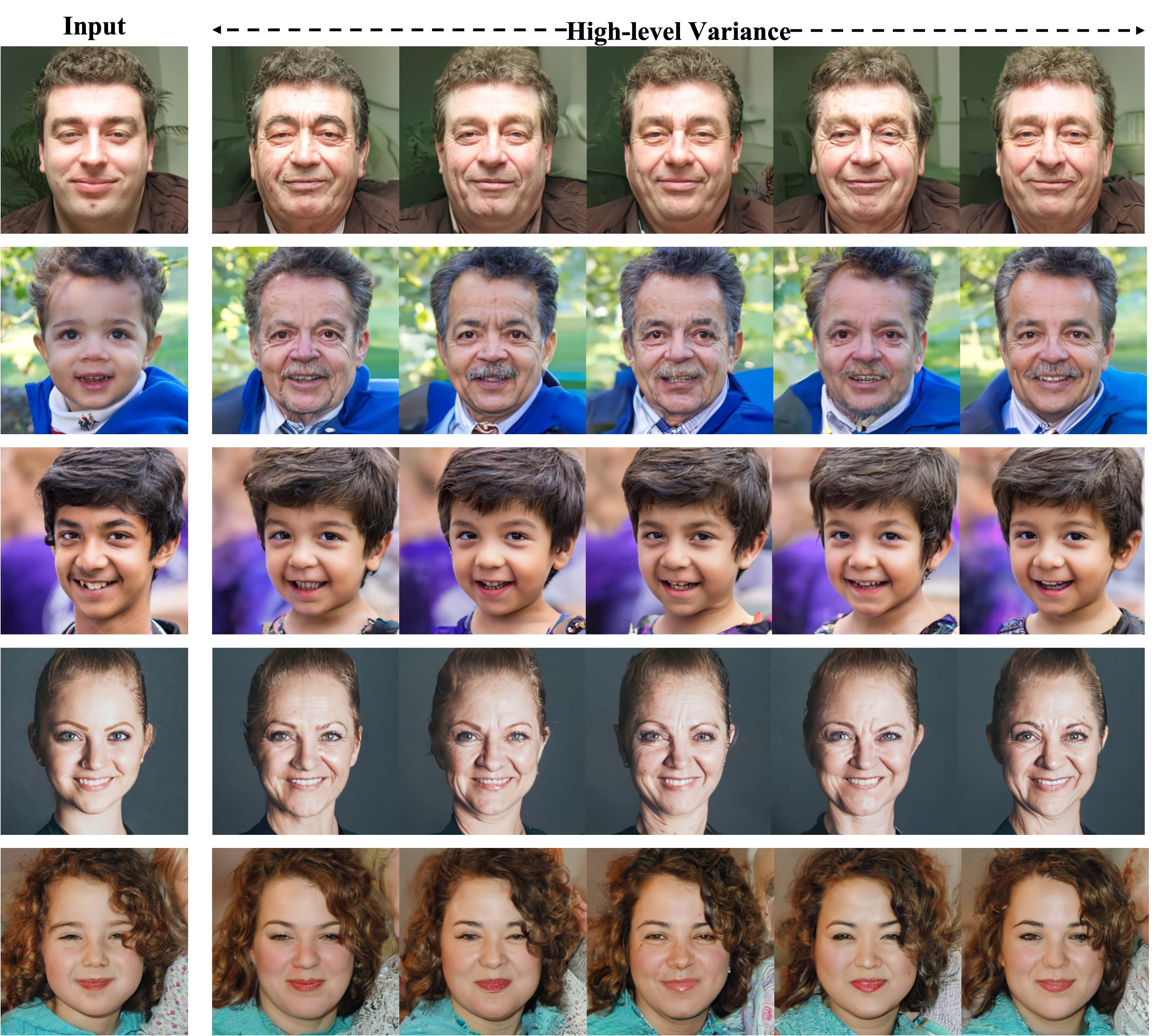}
\end{center}
   \caption{Pluralistic aging results with high-level variations on FFHQ-AT test set. }
\label{fig:5}
\end{figure*}

\begin{figure*}
\begin{center}
\includegraphics[width=1\linewidth]{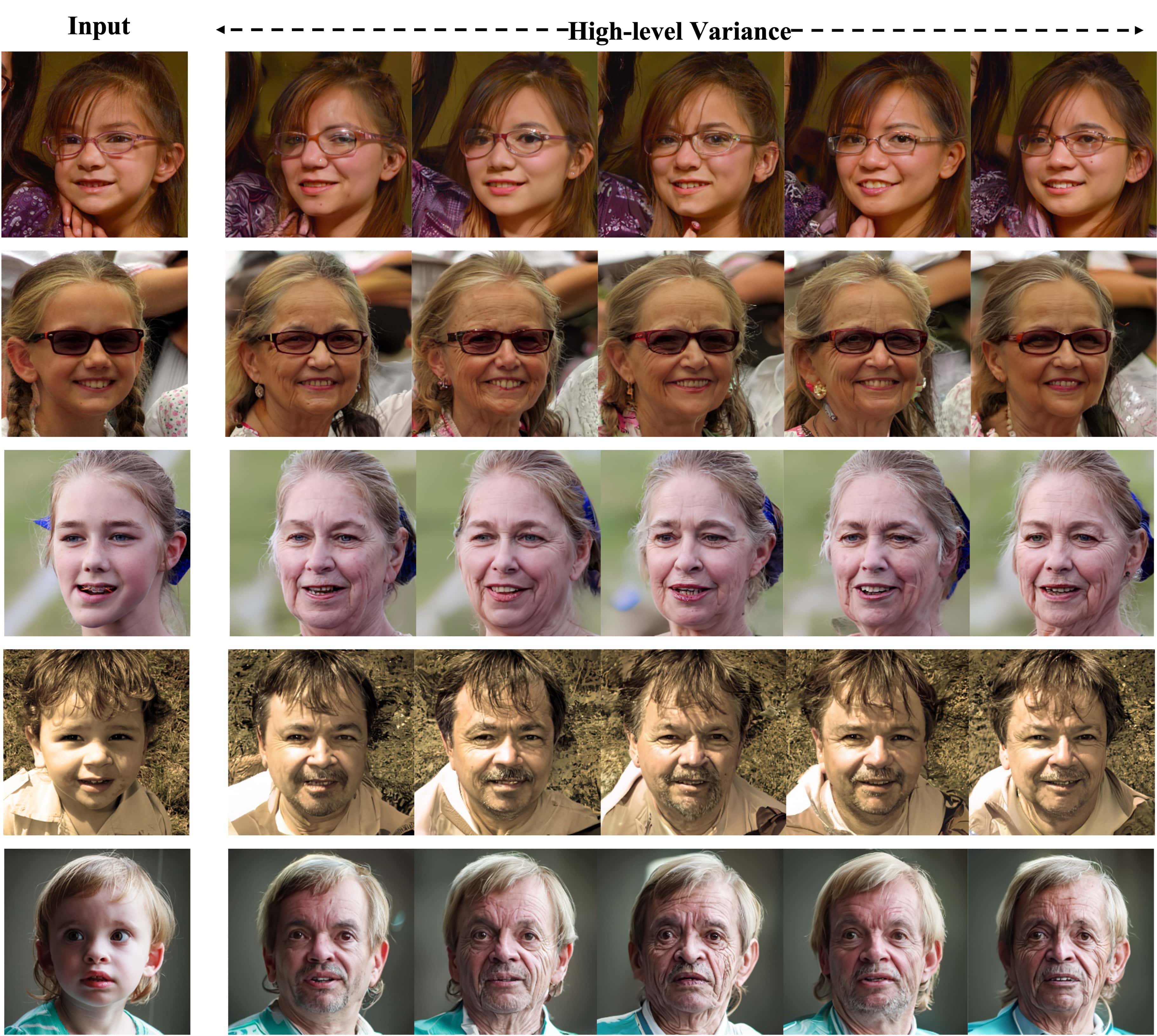}
\end{center}
   \caption{Pluralistic aging results with high-level variations on FFHQ-AT test set. }
\label{fig:6}
\end{figure*}

\begin{figure*}
\begin{center}
\includegraphics[width=1\linewidth]{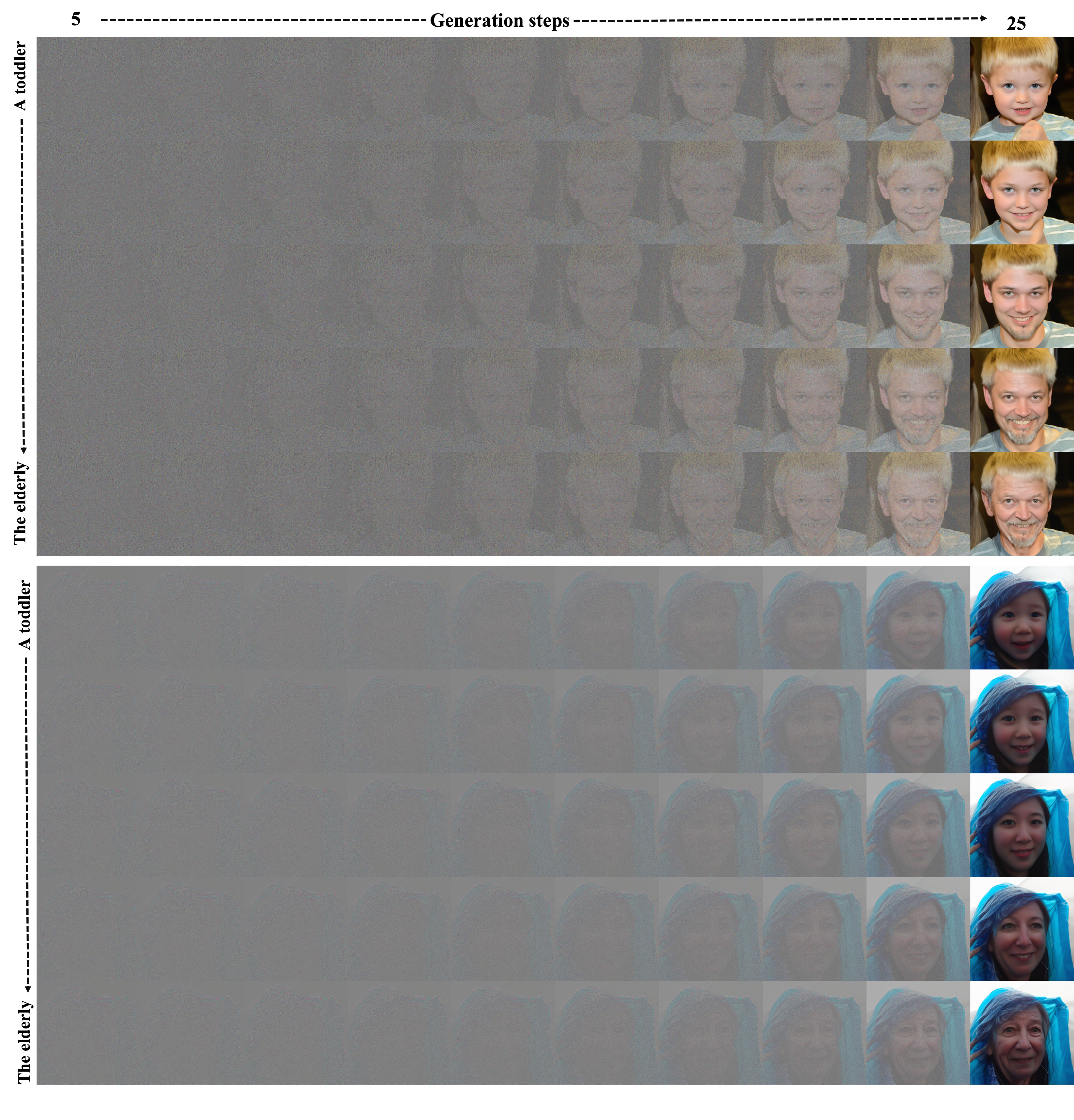}
\end{center}
   \caption{The intermediate generation results of diffusion decoder.}
\label{fig:inter}
\end{figure*}

We also show more reference-guided face aging results in Fig.~\ref{fig:3} and Fig.~\ref{fig:4}. Amazingly, our PADA can generate acne marks, which cannot be achieved by current face aging methods. The text-guided face aging results are shown in Fig.~\ref{fig:text}. More results based on the open-world age descriptions or arbitrary unseen facial images are shown in Fig.~\ref{fig:in_the_wild}. Although our PADA has not seen both these two variants during training, it still can generate plausible face aging results.
We show more diverse face aging results with high-level variations in Fig.~\ref{fig:5} and Fig.~\ref{fig:6}. The intermediate generation results of diffusion decoder are shown in Fig.~\ref{fig:inter}.


\end{document}